%% file: main.tex
\documentclass{article} 
\usepackage{colm2024_conference}

\usepackage{microtype}
\usepackage{hyperref}
\usepackage{url}
\usepackage{booktabs}

\usepackage{graphicx}
\usepackage{xcolor}
\usepackage{multirow}
\usepackage{hhline}
\usepackage{array}
\usepackage{hyperref}
\usepackage{CJK}

\usepackage{amsmath}
\usepackage{amssymb}
\usepackage{subcaption}

\usepackage{listings}
\usepackage{tabularx}

\usepackage{colortbl}
\definecolor{verylightblue}{rgb}{0.9, 0.95, 1}
\definecolor{lightblue}{rgb}{0.8,0.85,1}
\definecolor{mediumlightblue}{rgb}{0.7, 0.75, 1}

\title{Mistral-C2F: Coarse to Fine Actor for Analytical and Reasoning Enhancement in RLHF and Effective-Merged LLMs}

\author{\parbox{\linewidth}{
\centering{Chen Zheng  ~ Ke Sun ~ Xun Zhou 
} \\
{\rm Bytedance} \\
\texttt{\{chen.zheng1,ke.sun1,zhouxun\}@bytedance.com} \\
}
}

\colmfinalcopy 
\begin{document}

\maketitle

\begin{abstract}
Despite the advances in Large Language Models (LLMs), exemplified by models like GPT-4 and Claude, smaller-scale LLMs such as Llama and Mistral often struggle with generating in-depth and coherent dialogues. This paper presents a novel two-step Coarse-to-Fine Actor model to address the inherent limitations in conversational and analytical capabilities of small-sized LLMs. Our approach begins with the Policy-based Coarse Actor, employing a technique we term "Continuous Maximization". The Coarse Actor establishes an enhanced, knowledge-rich pool adept at aligning with human preference styles in analysis and reasoning. Through the RLHF process, it employs Continuous Maximization, a strategy that dynamically and adaptively extends the output length limit, enabling the generation of more detailed and analytical content. Subsequently, the Fine Actor refines this analytical content, addressing the generation of excessively redundant information from the Coarse Actor. We introduce a "Knowledge Residue Merger" approach, refining the content from the Coarse Actor and merging it with an existing Instruction model to improve quality, correctness, and reduce redundancies. We applied our methodology to the popular Mistral model, creating \textbf{Mistral-C2F}, which has demonstrated exceptional performance across 11 general language tasks and the MT-Bench Dialogue task, outperforming similar-scale models and even larger models with 13B and 30B parameters. Our model has significantly improved conversational and analytical reasoning abilities.
\end{abstract}

\section{Introduction}

Large Language Models (LLMs) such as GPT-4~\cite{gpt4openai} and Claude have revolutionized the field of Natural Language Processing (NLP) by demonstrating exceptional abilities in understanding context, generating text, and performing a wide array of language tasks~\cite{wu2023comparative}. Remarkably, smaller-scale LLMs like Llama~\cite{Touvron2023LLaMAOA,Touvron2023Llama2O}, Alpaca~\cite{alpaca}, Vicuna~\cite{vicuna2023}, and Mistral~\cite{Jiang2023Mistral7,zheng2024balancing} have also shown impressive results, performing well on general benchmarks like MMLU~\cite{hendryckstest2021} and BBH~\cite{srivastava2022beyond}. Furthermore, their general capabilities can be enhanced through Supervised Fine-Tuning (SFT) and Reinforcement Learning from Human Feedback (RLHF) alignment~\cite{ouyang2022training}.

However, these smaller-scale LLMs exhibit noticeable deficiencies in conversational abilities, even in aligning with human preference in dialogue style~\cite{Zhai2023InvestigatingTC}. The generated responses from these models often lack quality and consistency with the desired conversational preference~\cite{zheng2024balancing}. A significant shortcoming is that these smaller-scale models often produce responses that lack depth and detail, struggle with maintaining coherence over extended dialogues, and sometimes fail to exhibit robust underlying reasons or the analytical reasoning process. These issues reduce their effectiveness in complex applications like in-depth conversation and technical analysis~\cite{Wang2022TwostageLF,zhang2023balancing}.

Although Supervised Fine-Tuning (SFT) can sometimes reduce hallucinations, the resulting responses are typically short and lack substantial analysis, which negatively influences subsequent RLHF training. Furthermore, these models are often trained on datasets that do not fully engage their linguistic and cognitive capabilities, resulting in suboptimal performance in tasks that require high-level reasoning and detailed analysis~\cite{shen2024rethinking}.

In response to these challenges, we propose a novel two-step Coarse to Fine Analytical and Reasoning Enhancement LLM approach. ``Coarse to Fine'' (C2F) is a strategy used in various fields such as Question Answering for Long Document in NLP~\cite{choi-etal-2017-coarse}, and Semantic Matching in Information Retrieval~\cite{zheng2019rltm}, starting with broad, general answers and progressively refining to more detailed and specific answers that were overlooked in the coarse phase. However, the goal of our proposed LLM-based coarse to fine is completely different from traditional applications. The goal of our novel Coarse to Fine LLM is to transform a base LLM, which currently lacks integrated analytical and reasoning capabilities, into a RLHF LLM that aligns with human preferences and significantly enhances its analytical and reasoning abilities. To our best knowledge, this is the first time the ``Coarse to Fine'' approach introduced into LLMs alignment, marking a novel advancement LLMs field.

\textbf{Our first step is the ``Coarse Actor'' Analytical and Reasoning LLM.} In particular, we introduce the concept of "Continuous Maximization" (CM) in the direct application of RLHF to the base model. Continuous Maximization is a training strategy that dynamically and adaptively extends the output length limit, based on the model's sampling performance metrics, such as reward model scores and critic value function losses, rather than fixed iteration steps. This methodology prevents automatic stopping and enables the generation of more detailed and analytical content. In this paper, we design the Actor model to leverage this strategy, thereby enhancing its analytical and reasoning capabilities and creating a more analytical, in-depth, and knowledge-rich content.

However, the Coarse Actor cannot be used directly for responses or answers because serves as a knowledge-rich ``pool'', designed to enhance analytical and reasoning abilities. Extending the output to its length limits often results in excessive redundant information, as the model continues generating similar text without adequate termination.
To address this issue, \textbf{we introduce the second step: the ``Fine Actor'' Knowledge Refining LLM.} After the output from the Coarse Actor is generated, it is merged with existing Instruction model through new strategy 'Knowledge Residue Merger'. This merger allows the Coarse Actor to integrate its detailed analytical reasoning into the existing SFT model. Consequently, the Coarse-to-Fine actor approach retains the inherent advantages of the existing SFT model while significantly enhancing its analytical and reasoning capabilities in dialogue.

We applied our methodology to Mistral, the most popular base model, thereby creating \textbf{Mistral-C2F}. Our Mistral-C2F LLM has demonstrated exceptional performance across a spectrum of 11 general language tasks, surpassing benchmarks set by similar-scale models and even outperforming larger models such as those with 13B and 30B parameters. Through rigorous evaluating, including evaluations on the MT-BENCH benchmark, Mistral-C2F has shown to significantly excel in conversational ability and analytical reasoning, establishing SOTA on similar-scale and large-scale LLM performance.

To conclude, the contributions of our paper are:
\begin{enumerate}
  \item We introduced a novel two-step Coarse-to-Fine Actor LLM in PPO-based RLHF. In the Coarse Actor, we propose Continuous Maximization to develop a knowledge-rich pool, enhancing the model's analytical and reasoning capabilities in alignment with human preferences. This method significantly improves the model's ability to generate comprehensive and analytical content.
  \item In the Fine Actor, we introduced a Knowledge Residue Merger, refining the extensive content produced by the Coarse Actor, thereby improving quality and reducing redundancies in the generated text.
  \item Our enhanced LLM model, Mistral-C2F, has achieved SOTA performance across 11 benchmarks. We open-source Mistral-C2F on HuggingFace\footnote{Our Mistral-C2F is
available at \url{https://huggingface.co/zhengchenphd/Mistral-C2F-7B}.}, allowing the LLM community to delve deeper into our framework and facilitating further research.
\end{enumerate}

\begin{figure*}
\centering
\includegraphics[width=\textwidth,height=170pt]{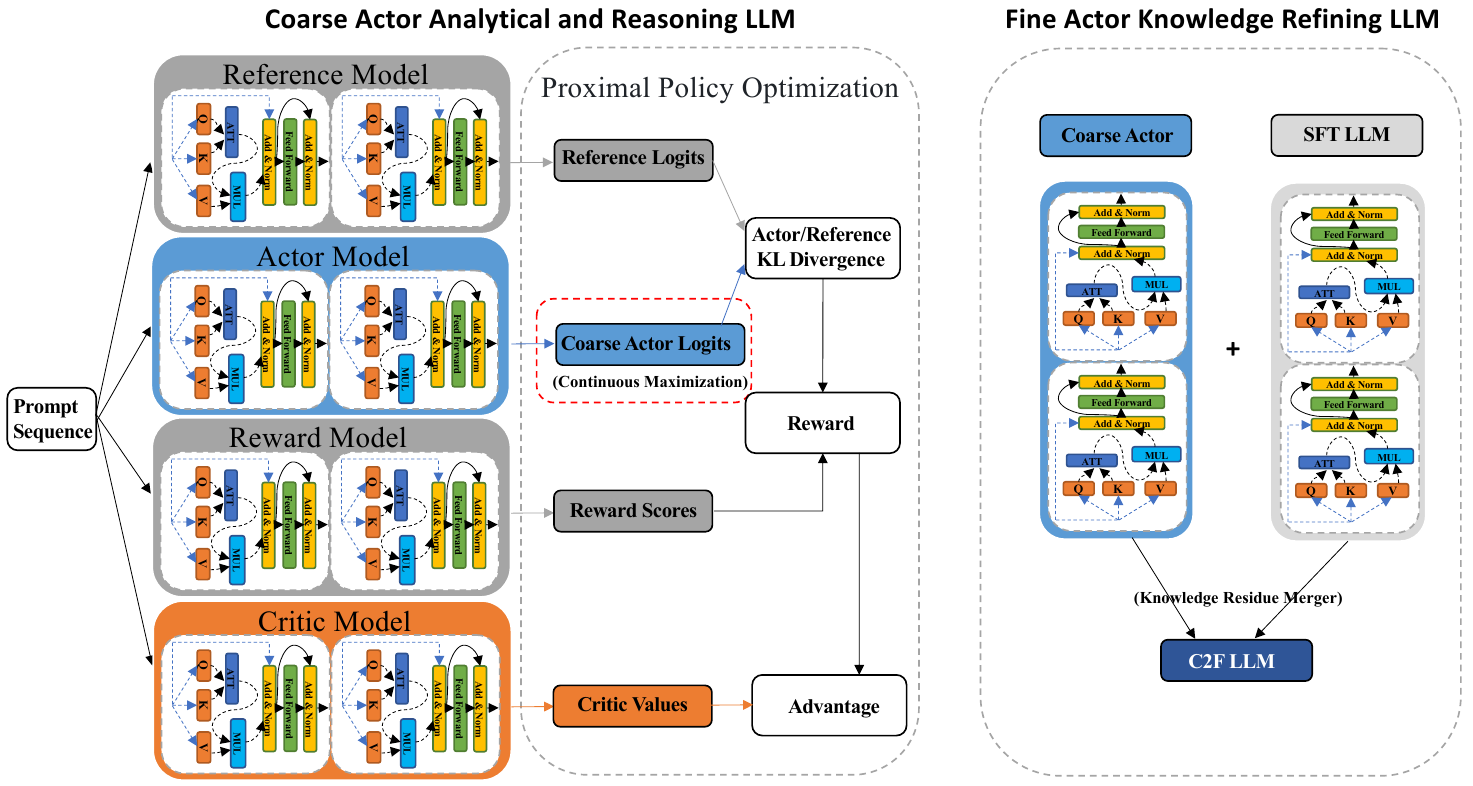}
\caption{Our Coarse to Fine Analytical and Reasonnig Enhancement LLM is composed of Coarse Actor Analytical and Reasoning LLM and Fine Actor Knowledge Refining LLM.\label{fig:architecture}}
\end{figure*}

\section{Related Works}

\subsection{LLM Alignment}

LLM Alignment allows the model to iteratively improve its performance by understanding and incorporating human preferences and nuances~\cite{ouyang2022training}, leading to more natural and engaging conversational abilities. 
Given the prevalent issues~\cite{Zheng2024ICEGRTIC} in language model outputs, there has been a shift towards fine-tuning to ensure they align more closely with human ethical standards. This alignment is primarily achieved through a method known as Reinforcement Learning from Human Feedback (RLHF)~\cite{ouyang2022training}.
The RLHF methodology initiates with the development of a reward function, which is shaped by a dataset embodying human preferences. This is followed by the fine-tuning of LLMs, aiming to optimize this reward. The optimization process leverages policy gradient methods, notably Reinforce \cite{williams1992simple} and Proximal Policy Optimization (PPO) \cite{schulman2017proximal}.

Recent advancements have led to the development of methods that directly utilize preference datasets, thereby eliminating the requirement for a predefined reward function. An example of this innovative approach is Direct Preference Optimization (DPO) \cite{rafailov2023direct}, which forges a direct link between the policy's logits and an implicit reward structure, facilitating the refinement of LLMs through a logistic loss function. This paradigm has been expanded through various methodologies, such as $\Phi$PO \cite{azar2023general}, KTO \cite{ethayarajh2024kto}, and Listwise Preference Optimization~\cite{liu2024lipo}.


\section{Model Description}


\subsection{Background}

In the context of Proximal Policy Optimization (PPO) based RLHF for an NLP chatbot, the actor-critic algorithm helps to optimize the chatbot's responses through iterative learning, where the action proposes responses and the critic evaluates their effectiveness. 
The PPO based RLHF approach for NLP chatbots aims to improve conversational abilities by learning from structured feedback, refining the policy in alignment with human preferences.

\textbf{Actor-Critic Algorithm in PPO based RLHF:}
The actor-critic algorithm is a type of RL method that combines the policy-based approach (actor) with the value-based approach (critic)~\cite{ouyang2022training}. The actor determines the action to take based on the current policy, while the critic evaluates the action by computing the value function, typically as a Temporal Difference error. The actor, parameterized by $\theta$, updates the policy based on the gradient of the expected reward. The critic, parameterized by $\phi$, estimates the value function, providing feedback to the actor for policy improvement.

\begin{equation}
\text{Actor: } \Delta\theta = \alpha \nabla_\theta \log \pi_\theta(a_t|s_t) A^\pi(s_t, a_t)
\end{equation}
\begin{equation}
\text{Critic: } \Delta\phi = \beta \partial\frac{(R_t - V_\phi(s_t))^2}{\partial \phi}
\end{equation}

where $\pi_\theta(a_t|s_t)$ is the policy, $A^\pi(s_t, a_t)$ is the advantage function, $R_t$ is the reward at time $t$, and $V_\phi(s_t)$ is the value function estimated by the critic.

\textbf{Reference Model and Reward Model in PPO based RLHF:}
The reference model in RLHF is used to guide the feedback process, ensuring that the chatbot's responses align with human preferences. The reward model, often derived from human feedback, quantifies the quality of the chatbot's responses, serving as the basis for training the policy and value functions.
In PPO, the objective is to maximize a clipped version of the policy’s advantage~\cite{ouyang2022training}, which prevents large policy updates and ensures stable training:

\begin{equation}
L^{CLIP}(\theta) = \mathbb{E}\left[\min(r_t(\theta) \hat{A}_t, \text{clip}(r_t(\theta), 1-\epsilon, 1+\epsilon) \hat{A}_t)\right]
\end{equation}

where $r_t(\theta) = \frac{\pi_\theta(a_t|s_t)}{\pi_{\theta_{old}}(a_t|s_t)}$ is the probability ratio, showing how much more likely the action is under the current policy than the reference policy, and $\epsilon$ is a hyperparameter that controls the clipping range.

\subsection{Coarse to Fine Actor for Analytical and Reasoning in RLHF}

\subsubsection{Coarse Actor: Policy based Analytical and Reasoning LLM}

In the first phase of our novel two-step Coarse-to-Fine LLM Actor for Analytical and Reasoning Enhancement approach, the focus is on the "Coarse Actor". The primary goal of the Policy-based Coarse Actor is to evolve the knowledge-rich base of the LLM, which initially lacks integrated analytical and reasoning capabilities, into an enhanced knowledge actor that aligns with human preferences and exhibits significantly amplified analytical and reasoning faculties. 
Inspired by the work in \cite{zheng2024balancing}, our approach leverages a variant of RLHF applied directly to the base model to achieve more profound and intrinsic general knowledge-rich capabilities.

In the Coarse Actor, we introduce the Continuous Maximization (CM) strategy, which is characterized by a dynamic and adaptive increase in the output length limit of the model. This process is governed not by fixed iteration steps, but by the model's performance metrics, such as reward model scores and critic value function losses. Specifically, the output length limit adjusts dynamically based on these sampling actor performance indicators, allowing for a flexible and responsive enhancement of the model's analytical and reasoning capabilities.
The mathematical formulation of the Coarse Actor is calculated as follows:
\begin{align}
O^{(t)} &= \pi_{\theta_{coarse}}(a_t | s_t, l^{(t-1)}), \\
\Delta\theta_{coarse} &= \alpha \nabla_{\theta_{coarse}} \log O^{(t)} A^\pi(s_t, a_t), \\
l^{(t)} &= f_{CM}(R_t(s_t, a_t), V_\phi(s_t)).
\end{align}
where $O^{(t)}$ represents the probability of taking action $a_t$ at time step $t$, as determined by the policy $\pi_{\theta}$ given the current state $s_t$ and the previous output length $l^{(t-1)}$.
$\Delta\theta_{coarse}$ is the update to the policy parameters, computed as the gradient of the log policy, and the advantage function $A^\pi(s_t, a_t)$.
$l^{(t)}$ represents the new output length limit, adjusted dynamically based on the CM logistic regression function \(f_{CM}\) of the reward model score $R_t(s_t, a_t)$ and the critic value function $V_\phi(s_t)$.

During the training process, if the reward score stabilizes and the critic value function exhibits minimal fluctuation, the model’s output length is gradually increased. Conversely, if performance metrics indicate instability or degradation (e.g., significant oscillation in critic loss), the output length may remain the same output length. This adaptive mechanism ensures that the Coarse Actor optimally extends the model's analytical and reasoning capabilities, gradually and effectively enhancing its performance based on real-time feedback.

Besides the training process, implementing the Coarse Actor model's sample generation process requires a critical modification: the prevention of generating end-of-sequence tokens such as [EOS] or $</s>$. During the pretraining phase, base models rarely produce end-of-sequence tokens, as learning to terminate responses is often acquired during the SFT process, characterized by training with short prompts ending with end-of-sequence tokens. To allow the actor model to progressively extend its generation, all stop tokens were removed from the RLHF dataset. 
Additionally, a specific System Prompt was employed, instructing the model to generate a detailed analytical and reasoning response without producing any end-of-sentence token like $</s>$ or [EOS]. 
The System Prompt is shown as below:

\fbox{
    \parbox{\textwidth}{
        \textit{Below is an instruction that describes a task. Write a detailed analytical and reasoning response that appropriately completes the request, and don't generate any end of sentence tokens.}
    }
}

By adopting the CM methodology in the Policy-based Coarse Actor, our approach creates a detailed and expansive knowledge pool that significantly enhances the model's analytical and reasoning output, thus preparing it for the subsequent Fine Actor.

\subsubsection{Fine Actor: Knowledge Refining LLM}

Although the Coarse Actor establishes a comprehensive and analytically knowledge-rich pool, it cannot be used directly due to the generation of excessive redundant information, leading to repetitious phrases. To refine this extensive content and reduce redundancy, we proceed to the second actor, the 'Fine Actor' knowledge refining LLM approach.

In the Fine Actor, we address the issue of redundancy by merging the Coarse Actor model with an existing Instruction model, creating a form of ``Knowledge Residue Merger''. This process integrates the detailed analytical and reasoning capabilities developed in the Coarse Actor into the SFT model, enhancing the model’s ability to produce precise and relevant responses in dialogues.
Unlike~\cite{Kim2023solar}, which merges several versions of SFT to balance different in-domain tasks' effects, our goal is to refine the extensive, knowledge-rich pool and reduce redundancy inherited from the Coarse Actor.
Given that the Instruction model and the Coarse model share identical architecture, we combine their learning parameters, $\theta$, in a weighted manner. The merged model parameters are computed as follows:
\begin{equation}
    \theta_{\text{fine}} = \gamma \theta_{\text{coarse}} + (1 - \gamma) \theta_{\text{SFT}}
\end{equation}

where $\gamma$ is a float number ranging from 0 and 1, representing the proportion of the Coarse model's parameters to be integrated into the Fine Actor model. This parameter blending approach allows the Fine Actor to leverage the extensive knowledge base of the Coarse Actor while maintaining the precise and context-specific capabilities of the SFT model, thereby enhancing the overall analytical and reasoning capabilities in dialogue interactions.

\section{Experiments}

\subsection{Experimental Setup}
Our experiments were conducted using the PyTorch framework~\cite{Paszke2019PyTorchAI}. The training of our Mistral-C2F model utilized the AdamW optimizer and the Sigmoid Linear Unit (SiLU) activation function~\cite{Elfwing2017SigmoidWeightedLU}. We employed $16$ A$100$ GPUs within the DeepSpeed framework~\cite{Rasley2020DeepSpeedSO}. Our approach to training included the use of bfloat$16$ precision for both the training process and model storage. Careful selection of learning rates was crucial: we set the learning rate for the actor at \(5 \times 10^{-6}\) and for the critic at \(5 \times 10^{-7}\), with a clipping range of $0.2$. To maintain consistency in training, we kept the discount factor at $0.95$. The Mistral-C2F model will be made available to the HuggingFace after the paper is accepted for promoting collaborative research and innovation.

\subsection{Dataset}
In this study, we employed the high-quality, human-annotated dataset from Anthropic~\cite{Bai2022TrainingAH}, which we designate as the helpfulness and harmlessness dataset, for both our Reward Model and RLHF model. This dataset contains an extensive array of paired samples. Each pair features a "chosen" and a "rejected" response to specific prompts, determined by human evaluators. The helpfulness dataset is segmented into three tranches: (1) data derived from Context-Distilled 52B language models~\cite{Askell2021AGL}, (2) selections made through rejection sampling, specifically best-of-16 sampling from a 52B preference model, and (3) selections from an iterative "online" sampling process. For the harmlessness aspect~\cite{Bai2022TrainingAH} curated data by selecting the more harmful responses from the 52B language models to assemble potentially harmful content. The objective of the helpfulness dataset is to navigate conversations towards positive outcomes, whereas the harmlessness dataset is inclined to direct interactions towards negative situations.

\subsection{General Task Evaluation}

Our evaluation was conducted using the GPT-Fathom LLM evaluation framework \cite{Zheng2023GPTFathomBL}, focusing on tasks prevalent in the general public language tasks. This evaluation framework was adopted to measure the general task effectiveness of Mistral-C2F and to place it within the broader context of the LLMs. We compared it with most popular open-source LLMs such as LLaMa2~\cite{Touvron2023Llama2O}, Vicuna~\cite{vicuna2023}, DeepSeek~\cite{Bi2024DeepSeekLS}, ICE-GRT~\cite{Zheng2024ICEGRTIC}, Mistral~\cite{Jiang2023Mistral7}, Mistral-Plus~\cite{zheng2024balancing}, among others. The evaluation encompassed $11$ benchmarks spanning diverse domains, including language comprehension and reasoning, to thoroughly evaluate capabilities such as MMLU~\cite{hendryckstest2021}, AGIEval~\cite{Zhong2023AGIEvalAH}, BBH~\cite{srivastava2022beyond}, ARC~\cite{Xu2023LLMsAT}, HellaSWAG~\cite{Zellers2019HellaSwagCA}, RACE~\cite{Lai2017RACELR}, etc. To ensure the reliable, fair and unbiased comparison of our evaluation, we maintained the same parameters as GPT-Fathom. This involved employing comparable input formats, metrics for evaluation, and conditions.

\section{Results}

\begin{table*}[h!]
\centering
\tiny
\resizebox{\textwidth}{!}{
\begin{tabular}{|l|c|c|c|c|c|c|c|c|c|c|c|}
\hline
Model & \textbf{MMLU} & \textbf{AGIEval} & \textbf{BBH} & \textbf{AGIEval-ZH} & \textbf{ARC-E} & \textbf{ARC-C} & \textbf{HellaSWAG} & \textbf{Winogrande} & \textbf{RACE-M} & \textbf{RACE-H} & \textbf{GSM8K}  \\
 & \textbf{5-shot} & \textbf{few-shot} & \textbf{3-shot} & \textbf{few-shot} & \textbf{1-shot} & \textbf{1-shot} & \textbf{1-shot} & \textbf{1-shot} & \textbf{1-shot} & \textbf{1-shot} & \textbf{8-shot}  \\
\hline
\rowcolor{verylightblue}
LLaMA 7B & 24.66\% & 20.05\% & 33.48\% & 23.68\% & 30.01\% & 26.71\% & 24.58\% & 50.36\% & 26.74\% & 29.19\% & 13.80\%  \\
\hline
\rowcolor{verylightblue}
Llama2 7B & 40.91\% & 25.97\% & 38.21\% & 26.21\% & 62.37\% & 48.46\% & 25.39\% & 50.36\% & 45.75\% & 39.54\% & 17.51\%  \\
\hline
\rowcolor{verylightblue}
Vicuna 7B & 38.49\% & 22.71\% & 37.26\% & 27.00\% & 69.74\% & 46.33\% & 17.37\% & 49.80\% & 50.21\% & 46.83\% & 21.68\% \\
\hline
\rowcolor{verylightblue}
DeepSeek 7B & 40.84\%	& 29.62\%	& 41.84\%	& 28.79\%	& 66.92\%	& 49.74\%	& 23.19\%	& 51.62\%	& 50.49\%	& 51.83\%	& 16.07\% \\
\hline
\rowcolor{verylightblue}
LLaMA 13B & 38.42\% & 26.78\% & 38.28\% & 25.51\% & 67.63\% & 49.23\% & 28.90\% & 47.51\% & 52.23\% & 48.51\% & 18.42\%  \\
\hline
\rowcolor{verylightblue}
Llama2 13B & 49.57\% & 34.85\% & 45.89\% & 32.93\% & 76.52\% & 55.63\% & 37.17\% & 52.17\% & 57.73\% & 55.09\% & 28.66\%  \\
\hline
\rowcolor{verylightblue}
Vicuna 13B & 35.84\% & 28.68\% & 39.27\% & 30.33\% & 60.23\% & 40.96\% & 30.03\% & 45.84\% & 59.19\% & 60.69\% & 24.56\%  \\
\hline
\rowcolor{verylightblue}
ICE-GRT 13B & 55.33\% & 34.92\% & 49.78\% & 34.23\% & 87.58\% & \textbf{70.99\%} & 39.37\% & 53.04\% & 75.91\% & 71.64\% & 51.48\%  \\
\hline
\rowcolor{lightblue}
Mistral-Base 7B & 56.24\% &	38.13\%	& 53.49\% & 34.35\%	& 88.17\%	& 70.39\%	& 38.90\%	& 57.93\%	& 75.97\%	& 72.13\%	& 39.88\%  \\
\hline
\rowcolor{lightblue}
Mistral-Instruct 7B & 53.72\% &	37.13\% &	47.80\% &	33.91\% &	85.98\% &	72.35\% &	61.64\% &	57.62\% &	78.83\% &	73.67\% &	46.70\% \\
\hline
\rowcolor{lightblue}
Mistral-Plus 7B (RLHF) & 56.29\%	& 38.36\%	& 54.69\%	& \textbf{36.01\%}	& 87.88\%	& 70.48\%	& 49.67\%	& 58.25\%	& 76.81\%	& 72.56\%	& 45.41\%  \\
\hline
\rowcolor{lightblue}
Mistral-Plus-Fine (Merge) & 56.69\%	& 35.86\%	& 52.60\%	& 32.59\%	& 86.24\%	& 72.57\%	& 54.19\%	& 60.09\%	& 77.09\%	& 72.41\%	& 48.82\%  \\
\hline
\rowcolor{mediumlightblue}
\textbf{Mistral-C2F 7B (Ours)} & \textbf{58.13\%} &	\textbf{39.50\%} &	\textbf{57.38\%} &	35.50\% &	\textbf{88.22\%} &	\textbf{74.06\%} &	\textbf{56.45\%} &	\textbf{60.69\%} &	\textbf{80.78\%} &	\textbf{75.76\%}	& \textbf{52.62\%} \\
\hline
\rowcolor{verylightblue}
LLaMA 30B & 50.38\% & 34.87\% & 49.70\% & 30.68\% & 82.41\% & 60.67\% & 31.31\% & 51.30\% & 65.18\% & 64.18\% & 35.10\%  \\
\hline
\rowcolor{verylightblue}
Llama2-70B & 64.72\% & 43.99\% & 65.22\% & 39.52\% & 93.43\% & 79.61\% & 68.45\% & 69.69\% & 87.60\% & 85.13\% & 56.56\% \\
\hline
\end{tabular}
}
\caption{Evaluating Benchmark Performance of LLMs in 11 General Language Tasks.\label{tab:general_tasks_scores}}
\end{table*}

In Table~\ref{tab:general_tasks_scores}, we focus on the performance of our Mistral-C2F model across various popular language benchmarks. First, the Mistral models, identified by their deep blue background, consistently outperform the LLaMA series, which have a light blue background, in all tasks. For instance, the Mistral-Base achieves a 56.24\% on MMLU, surpassing similar-capacity LLaMA models and even LLaMA 30B model.

Second, although the Mistral-Instruct~\cite{Jiang2023Mistral7} showed improved dialogue capability, it experienced a slight decline in general capabilities compared to the base model. In the AGIEval task, the Mistral-Instruct scored 37.13\%, which is a decrease of 2.62\% from the Mistral-Base's 38.13\%.
Moreover, Mistral-Plus 7B (RLHF)~\cite{zheng2024balancing} is our strongest baseline, highlighting the effectiveness of RLHF. Mistral-Plus remains consistent with the base model in general tasks and exhibits significant dialogue improvement, outperforming the Instruct model in both general tasks and MT-Bench.
Our Mistral-C2F demonstrates significant advancements in both general tasks and dialogue capabilities. It scored SOTA on MMLU, BBH, AGIEval, ARC, HellaSWAG, Winogrande, RACE, GSM8K. Our model surpasses both the LLaMA and Mistral models, showcasing the efficacy of our Coarse to Fine Actor approach in enhancing general performance.

We not only selected Mistral-Plus, a Mistral model based on PPO RLHF as our strongest baseline, but we also took a step further by merging Mistral-Plus and Mistral-Instruct as Mistral-Plus-Fine. This attempt aimed to make a fairer comparison and to verify whether simple merging any RLHF and Instruct model would yield positive effects.
As shown in Table~\ref{tab:general_tasks_scores}, we evaluated the Mistral-Plus-Fine and found that a simple merge of RLHF and Instruct models can cause negative fluctuations in some general evaluation sets. For instance, compared to Mistral-Plus, the performance dropped by 2.5\% on AGIEval and by 2.09\% on BBH. However, there were also some positive gains on other general evaluation sets. For example, it improved by 3.41\% on the GSM-8K dataset and by 0.4\% on MMLU. Nonetheless, in terms of large number of model performance, Mistral-Plus-Fine still has a significant gap compared to our Mistral-C2F. On GSM-8K, our model outperformed Mistral-Plus-Fine model by 3.8\%, and on AGIEval, we achieved a 3.64\% improvement. This once again validates the effectiveness of our Coarse-to-Fine LLM General Capability. We will show more Analysis in Section~\ref{sec:conversational}.

\begin{table}[ht]
\centering
\tiny
\setlength{\tabcolsep}{2pt}
\begin{tabular}{lc}
\hline
\textbf{Model}  & \textbf{MT Bench} \\
\hline
Llama 2 7B          & 2.74              \\
Alpaca 7B          & 4.54              \\
Vicuna 7B           & 6.00              \\
Llama 2 7B Chat     & 6.27              \\
Mistral-Instruct 7B & 6.84              \\
Llama 2 13B         & 2.61              \\
Alpaca 13B          & 4.53              \\
Llama 2 13B Chat    & 6.65              \\
Vicuna 13B          & 6.39              \\
Mistral-Plus 7B     & 6.92              \\
\textbf{Mistral-C2F 7B}     & \textbf{7.29 }             \\

\hline
\end{tabular}
\caption{Comparing Conversational Abilities Across different LLMs on MT-Bench.}
\label{tab:results_chat}
\end{table}

As shown in Table~\ref{tab:results_chat}, the Mistral-C2F 7B model stands out with the highest score of 7.29, indicating superior dialogue capability among the models compared. The Mistral-Instruct model shows a great performance with a score of 6.84, highlighting its enhanced conversational abilities. Notably, Mistral-Plus also performs well, achieving a score of 6.92, which suggests that the RLHF technique has positively impacted its dialogue performance. This performance underscores the strength of the Mistral-C2F in handling dialogue-intensive tasks. More detailed Analysis will show in Section~\ref{sec:conversational}.

\begin{figure}[h]
    \centering
    \begin{subfigure}[b]{0.32\textwidth}
        \centering
        \includegraphics[width=\textwidth]{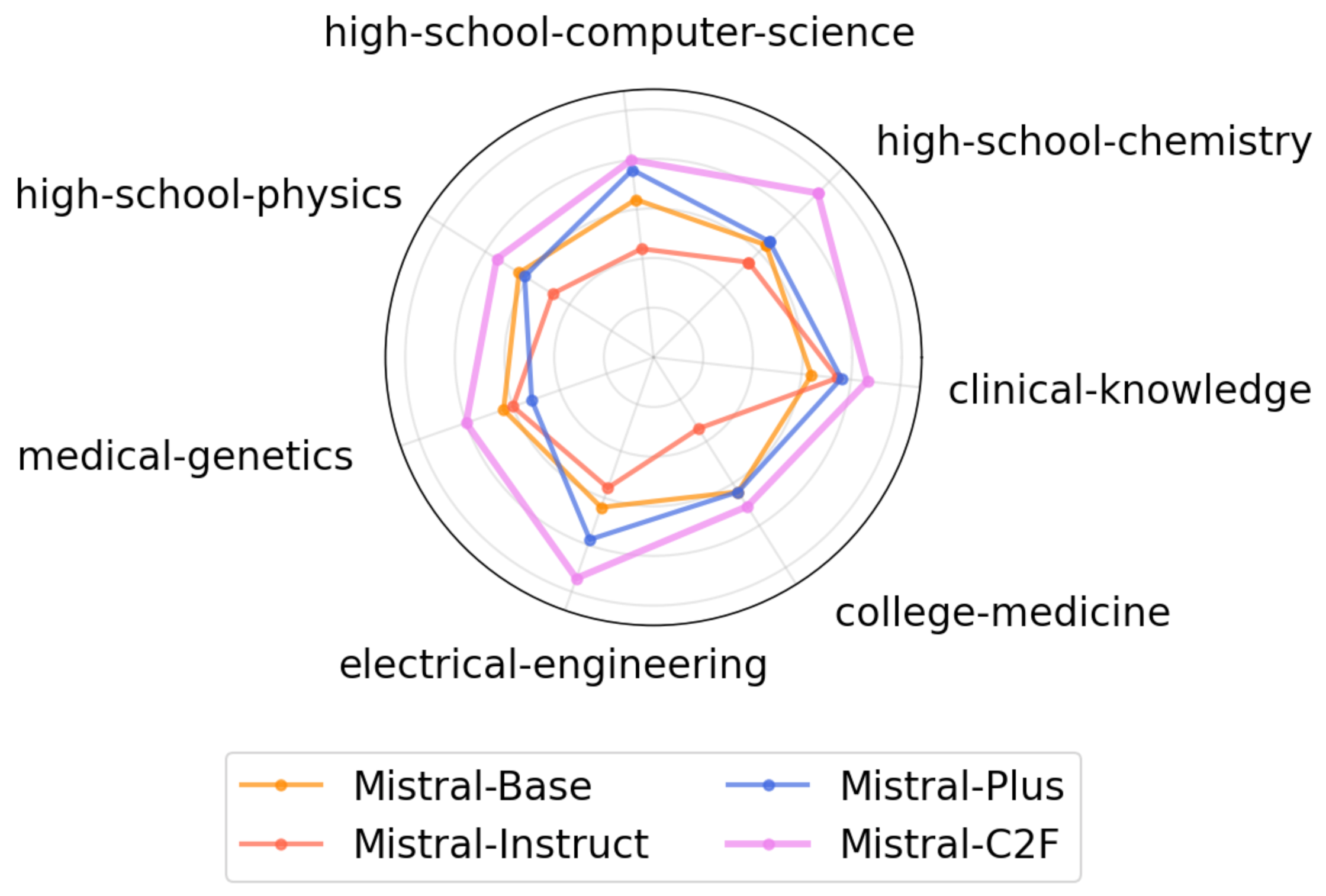}
        \caption{MMLU Performance Comparsion on Science Subjects.}
        \label{fig:mmlu_1}
    \end{subfigure}
    \hfill
    \begin{subfigure}[b]{0.32\textwidth}
        \centering
        \includegraphics[width=\textwidth]{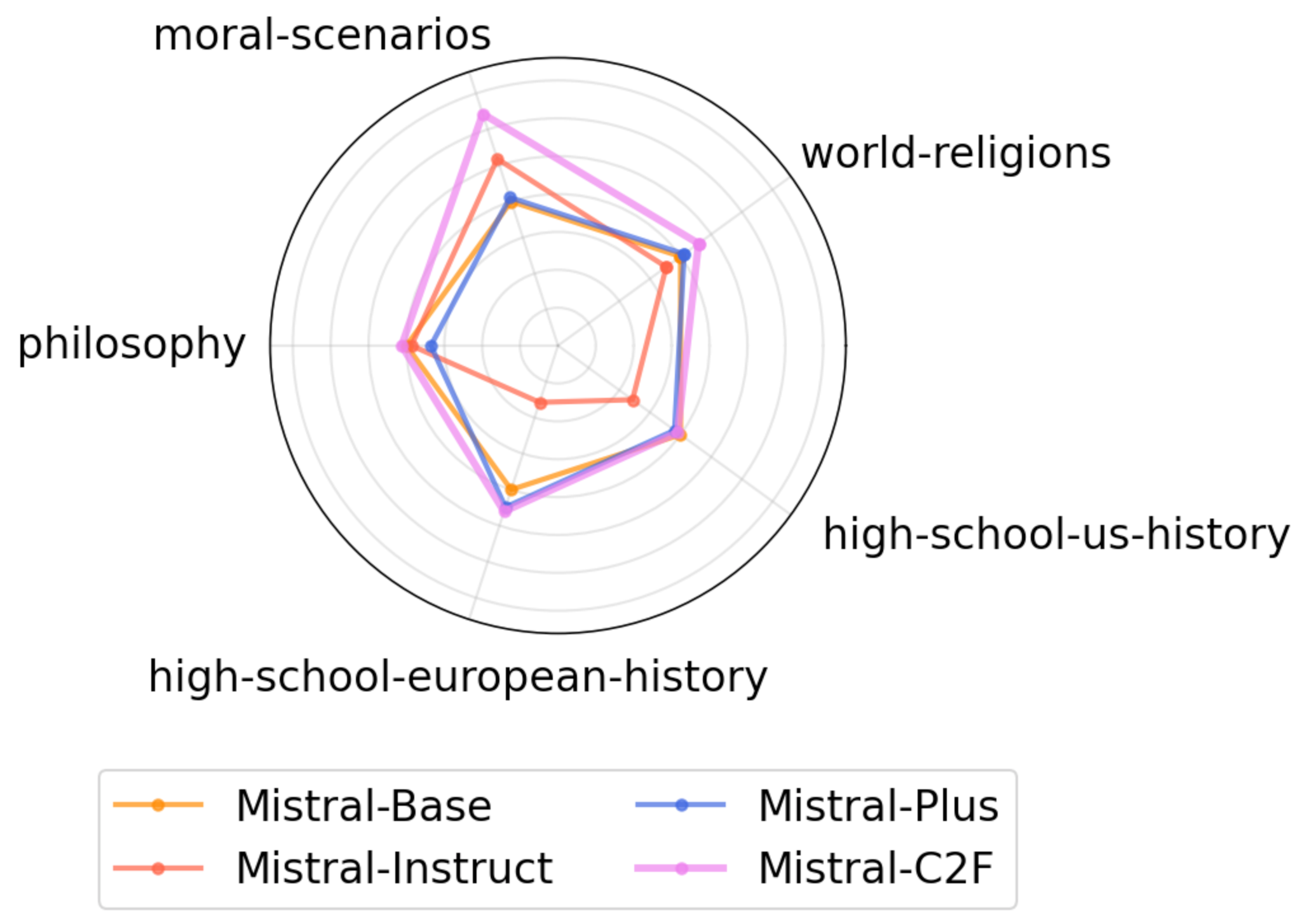}
        \caption{MMLU Performance Comparsion on Humanity Subjects.}
        \label{fig:mmlu_2}
    \end{subfigure}
    \hfill
    \begin{subfigure}[b]{0.32\textwidth}
        \centering
        \includegraphics[width=\textwidth]{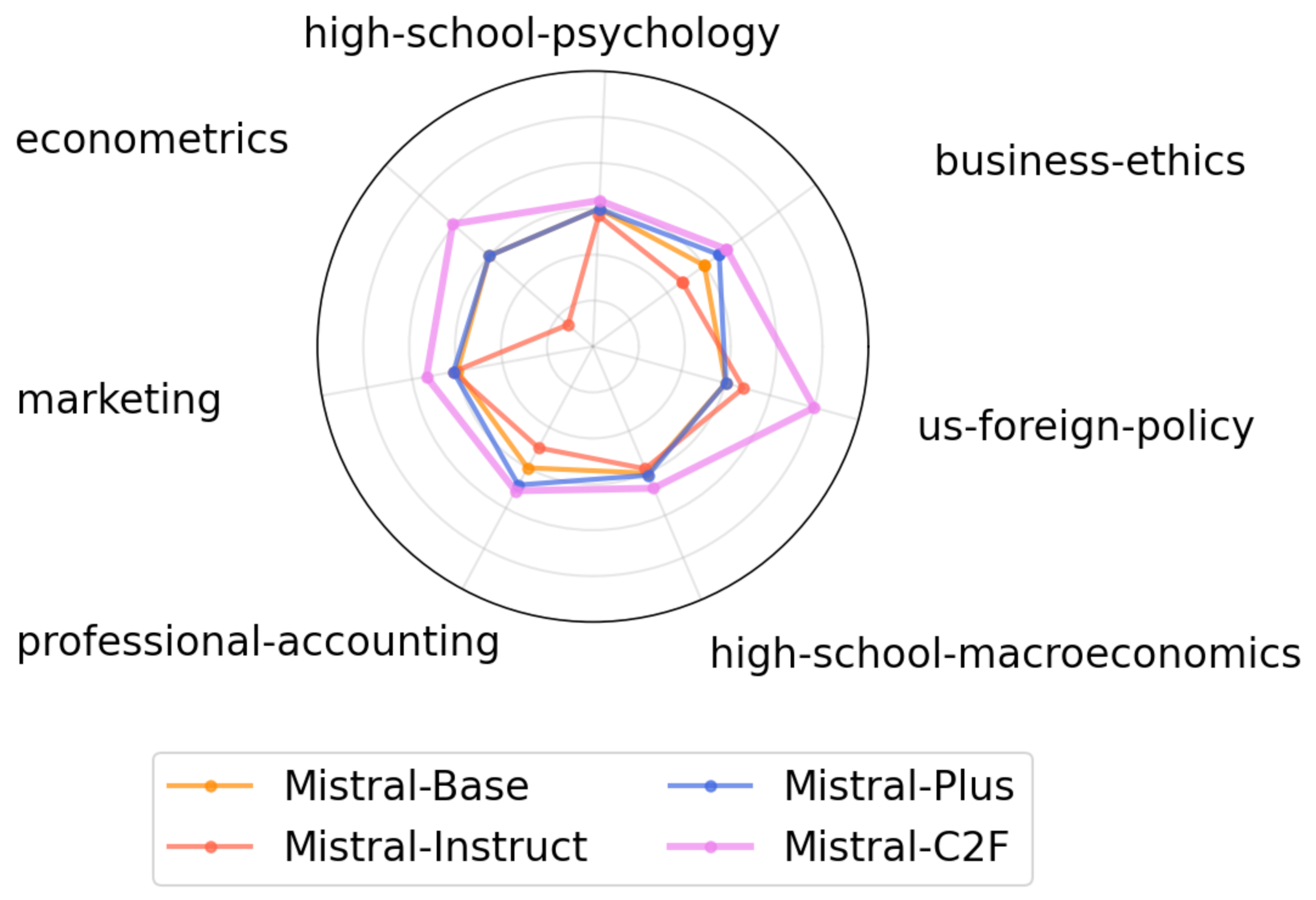}
        \caption{MMLU Performance Comparsion on Social Subjects.}
        \label{fig:mmlu_3}
    \end{subfigure}
    \caption{MMLU Score Comparsions between Mistral-C2F and Baseline LLMs.}
    \label{fig:radar}
\end{figure}

\section{Analysis}

\subsection{The Effectiveness of Coarse-to-Fine LLM General Capability}

The Massive Multitask Language Understanding (MMLU)~\cite{hendryckstest2021} is a significant benchmark in the field of natural language processing, featuring a wide array of 57 subjects. This extensive coverage makes MMLU a crucial dataset for evaluating the breadth and depth of a model's understanding across varied knowledge domains. These 57 subjects are categorized into three main groups: Science, Humanities, and Social, providing a comprehensive framework for assessing a model's proficiency in diverse academic fields.

In Figure~\ref{fig:radar}, we assess the performance of the Mistral family of LLMs across these categories. We observed that the Mistral-C2F model exhibits substantial improvements in the Science (Figure~\ref{fig:mmlu_1}) and Social (Figure~\ref{fig:mmlu_3}) categories, with notable gains in subjects like Chemistry, Electrical Engineering, Medical, Econometrics, and Policy. However, within the Humanities  (Figure~\ref{fig:mmlu_2}) category, significant improvement was primarily seen in the moral-scenarios subject, with modest enhancements in the remaining disciplines. This pattern underscores the Mistral-C2F model's strengths and areas for further refinement.

\subsection{The Effectiveness of Coarse Actor and Fine Actor}

\begin{table*}[h!]
\centering
\tiny
\resizebox{\textwidth}{!}{
\begin{tabular}{|l|c|c|c|c|c|c|c|c|c|c|c|}
\hline
Model & \textbf{MMLU} & \textbf{AGIEval} & \textbf{BBH} & \textbf{AGIEval-ZH} & \textbf{ARC-E} & \textbf{ARC-C} & \textbf{HellaSWAG} & \textbf{Winogrande} & \textbf{RACE-M} & \textbf{RACE-H} & \textbf{GSM8K}  \\
 & \textbf{5-shot} & \textbf{few-shot} & \textbf{3-shot} & \textbf{few-shot} & \textbf{1-shot} & \textbf{1-shot} & \textbf{1-shot} & \textbf{1-shot} & \textbf{1-shot} & \textbf{1-shot} & \textbf{8-shot}  \\
\hline
\rowcolor{verylightblue}
Mistral Instruct w/o Coarse Actor & 53.72\% &	37.13\% &	47.80\% &	33.91\% &	85.98\% &	72.35\% &	61.64\% &	57.62\% &	78.83\% &	73.67\% &	46.70\% \\
\hline
\rowcolor{lightblue}
Coarse Actor 0.1 + Intruct 0.9 &53.75\% &	31.86\%	& 38.18\%	& 31.67\%	& 83.75\%	&71.93\%	& 49.51\%	&59.98\%	&74.03\%	& 69.01\%	& 38.59\% \\
\hline
\rowcolor{lightblue}
Coarse Actor 0.2 + Intruct 0.8 &55.23\%	& 33.77\%	& 45.63\%	&32.44\%	&84.97\%	&73.29\%	&52.23\%	&60.85\%	&75.77\%	&71.38\%	&44.43\%\\
\hline
\rowcolor{lightblue}
Coarse Actor 0.3 + Intruct 0.7 &56.62\%	&35.69\%	&49.93\%	&32.96\%	&86.03\%	&74.32\%	&54.44\%	&60.93\%	&77.09\%	&73.10\%	&48.67\%\\
\hline
\rowcolor{lightblue}
Coarse Actor 0.4 + Intruct 0.6 &57.31\%	&36.55\%	&53.49\%	&34.30\%	&86.83\%	&74.32\%	&56.19\%	&60.54\%	&78.41\%	&74.27\%	&51.55\%\\
\hline
\rowcolor{lightblue}
Coarse Actor 0.5 + Intruct 0.5 &57.85\%	&37.21\%	&55.69\%	&34.26\%	&87.54\%	&74.66\%	&\textbf{57.67\%}	&60.62\%	&79.60\%	&75.53\%	&51.40\%\\
\hline
\rowcolor{lightblue}
Coarse Actor 0.6 + Intruct 0.4 &58.11\%	&38.88\%	&56.68\%	&34.80\%	&87.75\%	&\textbf{75.09\%}	&57.45\%	&\textbf{61.01\%}	&80.64\%	&75.50\%	&52.62\%\\
\hline
\rowcolor{lightblue}
\rowcolor{mediumlightblue}
\textbf{Coarse Actor 0.7 + Intruct 0.3} & \textbf{58.13\%} &	39.50\% &	\textbf{57.38\%} &	\textbf{35.50\%} &	88.22\% &	74.06\% &	56.45\% &	60.69\% &	\textbf{80.78\%} &	\textbf{75.76\%}	& \textbf{52.62\%} \\
\hline
\rowcolor{lightblue}
Coarse Actor 0.8 + Intruct 0.2 &57.96\%	&\textbf{39.70\%}	&56.68\%	&34.99\%	&88.30\%	&72.95\%	&53.33\%	&60.14\%	&80.15\%	&75.16\%	&52.16\%\\
\hline
\rowcolor{lightblue}
Coarse Actor 0.9 + Intruct 0.1 &57.36\%	&38.67\%	&55.56\%	&34.16\%	&\textbf{88.43\%}	&71.76\%	&45.23\%	&58.64\%	&78.20\%	&74.39\%	&48.29\%\\
\hline
\rowcolor{verylightblue}
Mistral Coarse Actor w/o Instruct & 56.34\%	&38.13\%	&53.60\%	&34.87\%	&88.17\%	&70.14\%	&28.72\%	&57.77\%	&76.11\%	&73.18\%	&45.56\%\\
\hline
\end{tabular}
}
\caption{Evaluating 11 Benchmark Performance of LLMs for Different Fine Actors. \label{tab:general_tasks_scores_ablation}}
\end{table*}

In our section, we conducted a robust Mistral-C2F ablation study across 11 datasets, ranging from pure Mistral-Instruct without Coarse Actor, through various combinations of Instruct and Coarse Actor with different knowledge residue mergers, to Mistral Coarse Actor without Instruct. We made two key observations: firstly, the Mistral Coarse Actor alone already showed a marked improvement over Mistral-Instruct, with approximately 3\% gain on MMLU and about 6\% on BBH, domonstrating the inherent strong capabilities. However, it tended to generate excessive redundant information, leading to suboptimal performance in conversation-oriented task. For example, in Table~\ref{table:eff_coarse}, the blue-colored text presents logical summaries of potential policy violations, and red text presents the detailed analysis. However, the brown text only redundantly reinforces blue color consequences without extra information. More examples are described in Table~\ref{table:eff_coarse_app} in Appendix.

\begin{table*}[h!]
\centering \tiny
\begin{tabular}{  m{3cm} | m{10cm} }
\hline
\textbf{Questions} & \textbf{Responses} \\
\hline
what ad policy does the following text violate?  lonely and thirsty for man & \textcolor{blue}{This text may violate several ad policies}, depending on the platform and the specific guidelines in place. \textcolor{blue}{\textbf{First}, the use of the word “lonely” may be considered a form of emotional manipulation}, which is often prohibited by ad policies. \textcolor{red}{This is because} advertisers are not allowed to use language or imagery that is intended to elicit feelings of loneliness or isolation in order to persuade people to buy their products or services. \textcolor{blue}{\textbf{Additionally}, the use of the word “thirsty” may be seen as a reference to sexual desire, which is often considered inappropriate in advertising.} Many ad policies prohibit the use of sexual innuendo or double entendres, as they can be seen as offensive or exploitative. \textcolor{blue}{\textbf{Furthermore}, the phrase “man” may be interpreted as a reference to a specific gender}, which can be considered discriminatory or exclusionary. \textcolor{red}{Advertisers are typically required to} use inclusive language that does not exclude or marginalize certain groups of people. \textcolor{blue}{Finally, the overall tone and message of the text, “lonely and thirsty for man,” may be perceived as negative and potentially harmful to the mental health of the audience.} \textcolor{red}{Ad policies often aim to} promote positive and uplifting messages in advertising, and content that suggests that a person’s loneliness or thirst for a specific gender is a problem to be solved may be seen as contributing to negative self-image or unrealistic expectations. \textcolor{blue}{\textbf{Overall}, it is important for advertisers to carefully consider the language and messaging they use in their ads to ensure that they are compliant with ad policies and respectful of their audience.} If the text violates any of the above-mentioned policies, it may be subject to removal or rejection by the platform or advertising agency. \textcolor{brown}{Additionally, advertisers may face penalties or sanctions for violating ad policies}, such as fines or loss of ad privileges. It’s always a good idea for advertisers to review and understand the ad policies of the platforms they are using and to seek guidance from legal or compliance experts if they have any questions or concerns about the appropriateness of their advertising content. By adhering to ad policies and promoting positive and inclusive messages, advertisers can help create a more respectful... (exceed maximum length)
\\
\hline
\end{tabular}
\caption{Case Study for Coarse Actor Generation.}
\label{table:eff_coarse}
\end{table*}

Through experimenting with different mergers of knowledge residue between the Instruct model and the Coarse Actor in Table~\ref{tab:general_tasks_scores_ablation}, we found that a 0.7 vs. 0.3 ratio achieved the best results, contrary to the anticipated 0.5 vs. 0.5 distribution. This surprising finding suggests that a larger proportion of the Coarse Actor is necessary, with the Instruct model serving as an auxiliary to enhance the quality of the generated responses, as described in Table~\ref{table:eff_fine}.

To further examine the effects of different Fine Actor mergers between the Instruct model and the Coarse Actor, we documented $9$ distinct responses generated from the same question in Table~\ref{table:eff_fine} and Appendix. These comparisons, described as ``compose an article which starts with flowers'', revealed that when the Coarse Actor's proportion was between 0.6 and 0.7, the model produced clear, logical, and well-structured responses, exhibiting a strong ability to conclude and analyze effectively in a segmented manner. However, starting from a 0.8 Coarse Actor ratio, the model began to lose its segmentation capability, with the answers gradually becoming less logical. The complete case studies from ratios of 0.1 to 0.9 are described in Appendix Table~\ref{table:eff_fine_2}.

\begin{table*}[h!]
\centering \tiny
\begin{tabular}{ m{1.5cm} |  m{1.5cm} | m{11cm} }
\hline
\textbf{Model} & \textbf{Question} & \textbf{Response} \\
\hline
Coarse Actor 0.6 + Intruct 0.4 & compose an article which starts with flowers. &  \textcolor{blue}{\textbf{Title: The Beauty of Flowers: A Symphony of Colors and Fragrances}}

Flowers have been a source of inspiration and delight for centuries. From the ancient Egyptians who used flowers in their burial rituals, to the Romantic poets who wrote about the beauty of flowers, these natural wonders have captured the imagination of people all over the world.

\textcolor{blue}{\textbf{The first thing that comes to mind when we think of flowers is their stunning beauty}}. With their vibrant colors and delicate petals, flowers are a feast for the eyes. Whether it's the bright red of a rose, the soft pink of a peony, or the cheerful yellow of a sunflower, each flower has its own unique beauty that never fails to captivate us.

But flowers are not just beautiful to look at. They also have a wonderful fragrance that can transport us to another world. From the sweet scent of a jasmine flower to the earthy aroma of a lily, each flower has its own distinctive fragrance that can evoke powerful emotions and memories.

\textcolor{blue}{\textbf{Flowers are also an important part of our culture and traditions.}}In many countries, flowers are used to express emotions and sentiments. For example, in the language of flowers, a red rose is a symbol of love and passion, while a white rose represents purity and innocence.

\textcolor{blue}{\textbf{In addition to their cultural significance, flowers also have practical uses.}} Many flowers are used in medicine to treat a variety of ailments. For example, chamomile flowers are often used to make tea that can help soothe an upset stomach or promote better sleep.

\textcolor{blue}{\textbf{In conclusion, flowers are truly a wonder of nature.}} With their stunning beauty, wonderful fragrance, cultural significance, and practical uses, flowers are a source of inspiration and delight for people all over the world. So the next time you see a beautiful flower, take a moment to appreciate its beauty and wonder.
\\
\hline
Coarse Actor 0.8 + Intruct 0.2 & compose an article which starts with flowers.
&  Flowers are a beautiful and beloved part of nature. They come in a wide variety of colors, shapes, and sizes, and they have been admired and appreciated by humans for centuries. Flowers have a special ability to brighten up any space and bring a sense of joy and happiness to those who see them. They are often used as symbols of love, friendship, and celebration, and they are a popular choice for gifts and decorations. Flowers also play an important role in the ecosystem, providing food and shelter for a variety of insects and other animals. They help to purify the air and contribute to the overall health and well-being of the environment. In addition to their aesthetic and ecological value, flowers also have a long history of use in traditional medicine. Many flowers contain compounds that have been shown to have therapeutic properties, and they have been used to treat a wide range of ailments, from headaches and colds to more serious conditions like cancer and heart disease. \textcolor{blue}{Overall, flowers are a truly remarkable and multifaceted part of our world.} They bring beauty and joy to our lives, contribute to the health and well-being of the environment, and have a rich history of use in traditional medicine. Whether you're admiring a bouquet of freshly cut flowers, or simply enjoying the sight of a beautiful flowering plant in your garden, there's no denying the power and magic of these wonderful creations of nature.\\
\hline
\end{tabular}
\caption{Case Study in Fine Actor Generation task for the Flower Story.}
\label{table:eff_fine}
\end{table*}

\subsection{The Effectiveness of Conversational Ability}
\label{sec:conversational}

\begin{figure}[h]
    \centering
    \begin{subfigure}[b]{0.4\textwidth}
        \centering
        \includegraphics[width=\textwidth]{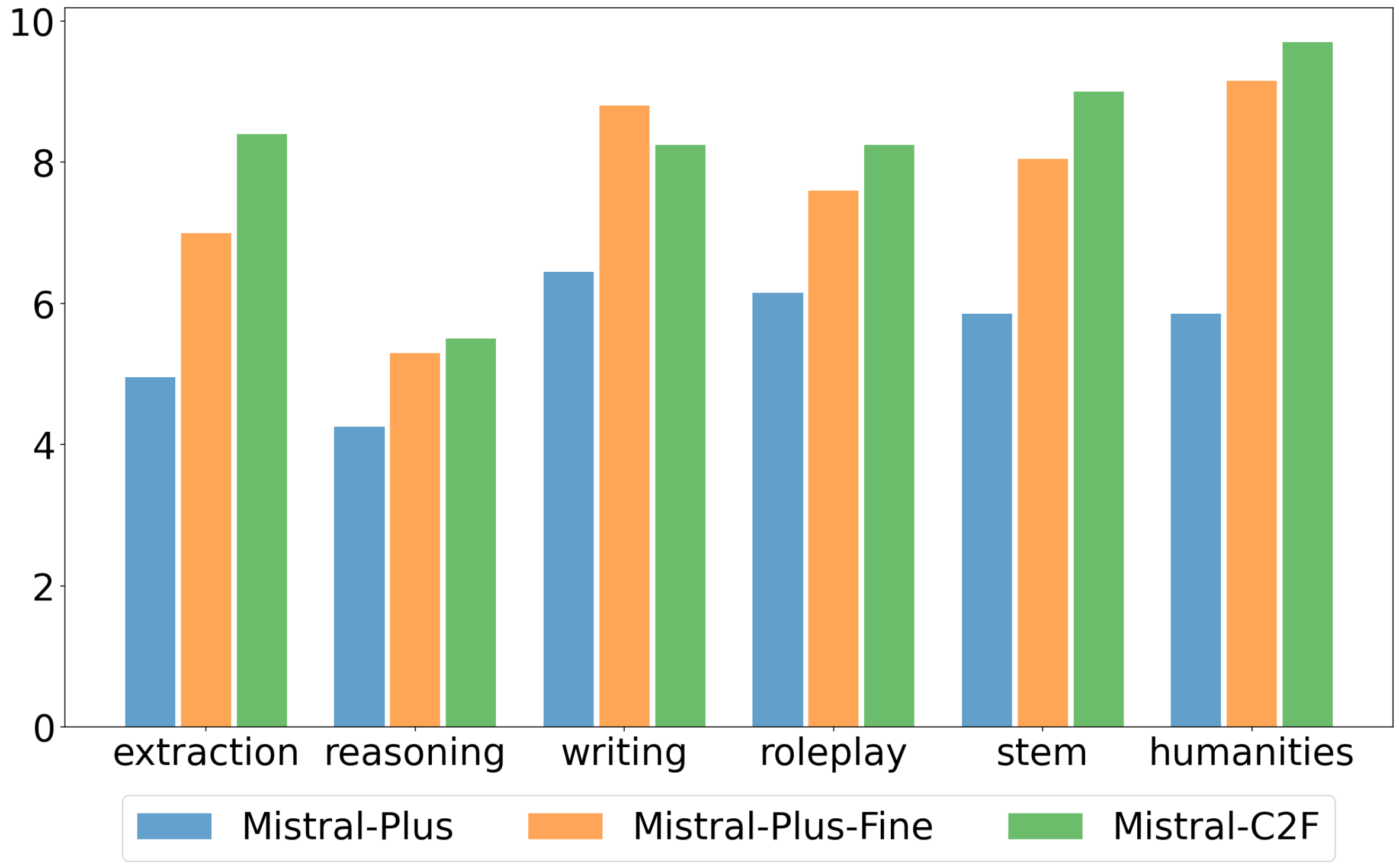}
        \caption{MT-Bench Performance Comparsion on 6 different Categories.}
        \label{fig:mt_bench_ana1}
    \end{subfigure}
    \hfill
    \begin{subfigure}[b]{0.25\textwidth}
        \centering
        \includegraphics[width=\textwidth]{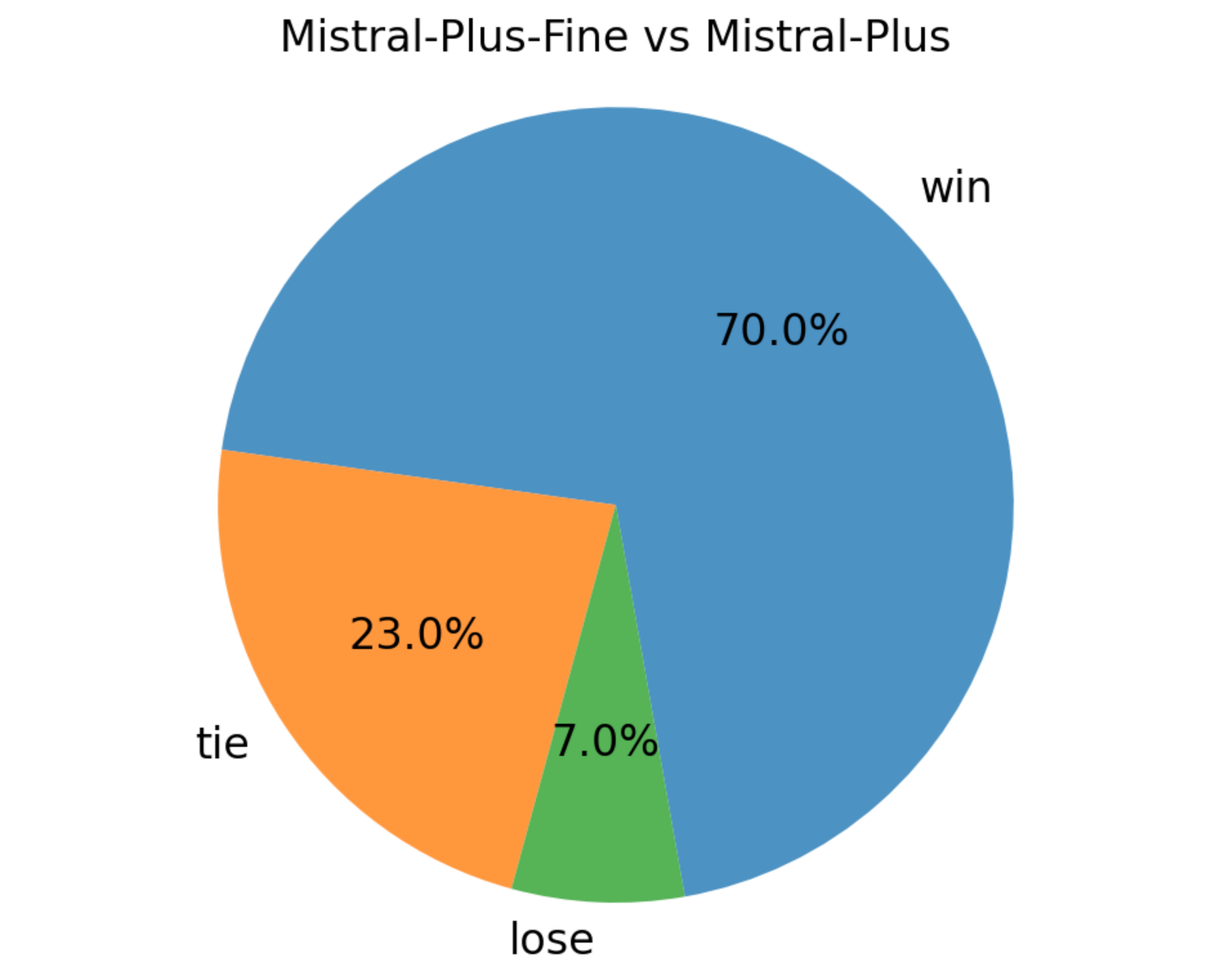}
        \caption{Win Rate Comparsion between Mistral-Plus and Mistral-Plus-Fine.}
        \label{fig:win_ana1}
    \end{subfigure}
    \hfill
    \begin{subfigure}[b]{0.25\textwidth}
        \centering
        \includegraphics[width=\textwidth]{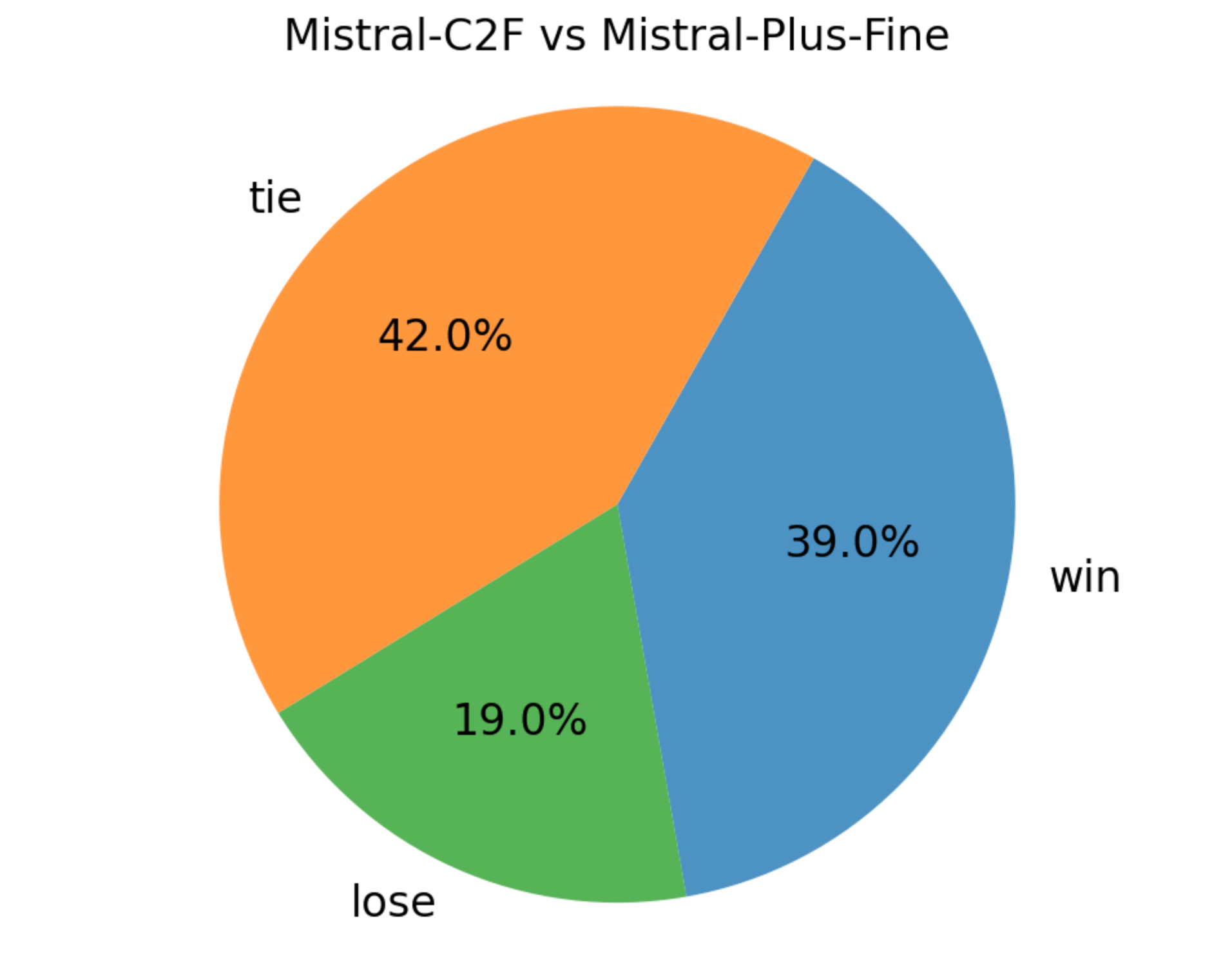}
        \caption{Win Rate Comparsion between Mistral-Plus-Fine and Mistral-C2F.}
        \label{fig:win_ana2}
    \end{subfigure}
    \caption{Six Categories MT-Bench and Win Rate between Baselines and Mistral-C2F.}
    \label{fig:mt_bench_comp}
\end{figure}

Following by MT-Bench Judgement~\cite{zheng2024judging}, we evaluated the conversational abilities of the Mistral-C2F model. Figure~\ref{fig:mt_bench_ana1} illustrates the 6 categories within the MT-Bench tasks, such as extraction, reasoning, writing, roleplay, stem, and humanities. We compared the performance against two strong baselines: Mistral-Plus and 
our proposed Mistral-Plus-Fine, obtained through Mistral-Plus and Mistral-Instruct Fine Actor. We observe that the Mistral-C2F model achieved SOTA results in five out of these six categories, confirming the effectiveness of our model. In the Appendix Table~\ref{table:eff_mt_app}, we present a selection of cases as a case study, demonstrating significant improvements in dialogue quality, logical consistency, and analytical depth, thereby proving the value of the Coarse-to-Fine approach.

In Figure~\ref{fig:win_ana1} and Figure~\ref{fig:win_ana2}, we calculated the win rates between our Mistral-C2F model and the two strongest baseline models. From Figure~\ref{fig:win_ana1}, it is evident that the Mistral-Plus model, after integration with our proposed Mistral-Plus-Fine actor, 
significantly outperforming the Mistral-Plus model. 
Moreover, Figure~\ref{fig:win_ana2} reveals that the win rate for the Mistral-C2F model reached 39\%, substantially higher than the 19\% of the more enhanced Mistral-Plus-Fine.

\subsection{Case Study Between Baseline Models and Our Mistral-C2F}

To further domonstrate the efficacy and efficiency of Mistral-C2F model in both general and conversational tasks, we conducted extensive case study in Table~\ref{table:eff_general_dia_1}, Table~\ref{table:eff_general_dia}, Table~\ref{table:eff_general_dia_instruct_and_rl}, and Table~\ref{table:eff_mt_app}. Each case was comparatively analyzed across three model variants: Mistral-Plus, Mistral-Plus-Fine, and Mistral-C2F, with detailed comparisons presented in the Appendix Section. 
Through these comprehensive case analyses, we consistently observed that the responses generated by the Mistral-C2F model exhibited significant improvements in various aspects. 
Whether it was writing, reasoning, or analytical tasks, the quality of the generated content, the logical coherence of the responses, and the length of the output all showed notable enhancements with the Mistral-C2F.

\begin{table*}[h!]
\centering \tiny
\resizebox{1.\linewidth}{!}{
\renewcommand{\arraystretch}{1.5}
\begin{tabular}{  m{3cm} | m{3cm} | m{3cm} | m{5cm} }
\hline
\textbf{Questions} & \textbf{Mistral Plus} & \textbf{Mistral-Plus-Fine} & \textbf{Mistral-C2F} \\
\hline
You're meeting a friend for coffee, but they're running late. How would you greet them when they arrive?

& 1. Smile and say, "Hi, sorry you're late!" \newline 2. Offer them a seat and ask how their day is going. \newline 3. Be understanding and patient with them. \newline 4. Let them know that it's no problem and that you're glad to see them.

& 1. Smile and greet them warmly, "Hi, it's great to see you!" \newline 2. Offer them a seat and ask if they'd like anything to drink. \newline 3. Express understanding and empathy for their lateness, "No worries, I understand that things come up."

&  1. Smile and say, "Hi! No worries, take your time. I'm just enjoying the coffee and the atmosphere." \newline 2. Offer a warm greeting, such as, "Hey there! Don't let my smile fool you, I'm just happy to see you. Take your time, I'm in no rush." \newline 3. If you're in a public place, you could say, "Hey, no problem. I'm just enjoying the coffee and the people-watching. Take your time, I'll be here when you're ready." \newline 4. If you're in a more private setting, such as a friend's home, you could say, "Hey, no worries. I'm just catching up on some reading. Take your time, I'll be here when you're ready." \newline The key is to be understanding and supportive, while also making it clear that you're happy to see them and that they're not keeping you waiting.
\\
\hline
How do you usually spend your weekends, and what activities do you enjoy the most?

&  I usually spend my weekends doing a variety of activities, including spending time with family and friends, going for walks, cooking, and reading. I especially enjoy spending time outdoors, whether it’s hiking, biking, or just enjoying the fresh air.

&  I usually spend my weekends catching up on household chores, running errands, and spending time with my family. I enjoy cooking, gardening, and going for walks with my dog. On occasion, I also like to go out with friends for dinner or a movie.

& I typically spend my weekends relaxing and engaging in activities that I enjoy. One of my favorite weekend activities is going for a hike in nature. I find it incredibly refreshing to be surrounded by the beauty of the natural world. Another activity I enjoy is cooking. I love trying out new recipes and experimenting with different flavors and ingredients. I also enjoy spending time with friends and family, whether it's going out for dinner or just hanging out at home. Overall, I try to make the most of my weekends by doing things that make me happy and help me relax and recharge for the week ahead.

\\
\hline
\end{tabular}
}
\caption{Case Study in General Dialogue.}
\label{table:eff_general_dia_1}
\end{table*}

\section{Conclusion}

This paper introduced the novel Coarse-to-Fine Actor model, an innovative approach designed to overcome the inherent limitations in conversational and analytical capabilities of existing LLMs. By the effectiveness of Coarse and Fine Actors, we addressed the crucial aspects of knowledge depth and quality in generated content.
The Coarse Actor, utilizing the Continuous Maximization technique, significantly enhanced the knowledge-rich pool and analytical reasoning of LLMs. 
The subsequent Fine Actor refined this extensive content through knowledge residue merger approach, merging it with an existing Instruction model to improve precision and eliminate redundancies. 
Our application of this methodology to our Mistral-C2F, which exhibited exceptional performance across a broad range of language tasks, outperforming similar-scale and even larger models.

\section*{Acknowledgements}
We are sincerely grateful to Ruoqi Zhang, Wumo Yan, and Yukun Ma for being early-stage users of our Mistral-C2F and for providing insightful case feedback that has significantly enriched our research. Additionally, we extend our appreciation to Hanzhi Zhou and Jay Yang for their engineering support in building key components of the infrastructure. Furthermore, we deeply appreciate Liang Xiang's support of our research project. We also thank the anonymous reviewers for their valuable suggestions.


\bibliography{colm2024_conference}
\bibliographystyle{colm2024_conference}

\newpage
\appendix
\section{Appendix}

\input{appendix}

\end{document}

%% file: appendix.tex
\begin{table*}[h!]
\centering \tiny
\resizebox{1.\linewidth}{!}{
\renewcommand{\arraystretch}{1.5}
\begin{tabular}{  m{3cm} | m{3cm} | m{3cm} | m{5cm} }
\hline
\textbf{Questions} & \textbf{Mistral Plus} & \textbf{Mistral-Plus-Fine} & \textbf{Mistral-C2F} \\
\hline
You're meeting a friend for coffee, but they're running late. How would you greet them when they arrive?

& 1. Smile and say, "Hi, sorry you're late!" \newline 2. Offer them a seat and ask how their day is going. \newline 3. Be understanding and patient with them. \newline 4. Let them know that it's no problem and that you're glad to see them.

& 1. Smile and greet them warmly, "Hi, it's great to see you!" \newline 2. Offer them a seat and ask if they'd like anything to drink. \newline 3. Express understanding and empathy for their lateness, "No worries, I understand that things come up."

&  1. Smile and say, "Hi! No worries, take your time. I'm just enjoying the coffee and the atmosphere." \newline 2. Offer a warm greeting, such as, "Hey there! Don't let my smile fool you, I'm just happy to see you. Take your time, I'm in no rush." \newline 3. If you're in a public place, you could say, "Hey, no problem. I'm just enjoying the coffee and the people-watching. Take your time, I'll be here when you're ready." \newline 4. If you're in a more private setting, such as a friend's home, you could say, "Hey, no worries. I'm just catching up on some reading. Take your time, I'll be here when you're ready." \newline The key is to be understanding and supportive, while also making it clear that you're happy to see them and that they're not keeping you waiting.
\\
\hline
How do you usually spend your weekends, and what activities do you enjoy the most?

&  I usually spend my weekends doing a variety of activities, including spending time with family and friends, going for walks, cooking, and reading. I especially enjoy spending time outdoors, whether it’s hiking, biking, or just enjoying the fresh air.

&  I usually spend my weekends catching up on household chores, running errands, and spending time with my family. I enjoy cooking, gardening, and going for walks with my dog. On occasion, I also like to go out with friends for dinner or a movie.

& I typically spend my weekends relaxing and engaging in activities that I enjoy. One of my favorite weekend activities is going for a hike in nature. I find it incredibly refreshing to be surrounded by the beauty of the natural world. Another activity I enjoy is cooking. I love trying out new recipes and experimenting with different flavors and ingredients. I also enjoy spending time with friends and family, whether it's going out for dinner or just hanging out at home. Overall, I try to make the most of my weekends by doing things that make me happy and help me relax and recharge for the week ahead.

\\
\hline
Write a short poem about the beauty of nature.

& The beauty of nature is a sight to behold,

With its vibrant colors and soothing sounds.

The trees sway in the breeze,

The flowers bloom with joy,

And the birds sing their sweet songs.

Nature is a gift that we should cherish and protect,

For it is a source of peace and tranquility.
& The beauty of nature,

A sight to behold,

The sunrise, the sunset,

The flowers, the trees, the birds,

All in perfect harmony,

A gift from the Creator,

A reminder of His love.
& In the heart of the forest,

Where the trees dance in the breeze,

Nature's beauty unfolds,

A symphony of colors and sounds.
\newline\newline
The sun paints the sky with hues,

As the moon whispers gentle lullabies,

The river sings a melody,

While the birds serenade the trees.
\newline\newline
In this enchanting world,

Nature's beauty holds us in awe,

A reminder of the wonders that surround us,

A testament to the magic that exists in every moment.
\\
\hline
Passage: "The Great Gatsby" by F. Scott Fitzgerald is a quintessential American novel that explores the concepts of decadence, idealism, and resistance to change. Set in the 1920s, it portrays the story of Jay Gatsby, a self-made millionaire, and his pursuit of Daisy Buchanan, a wealthy young woman whom he loved in his youth.
Question: What central theme does "The Great Gatsby" explore?
& The central theme of "The Great Gatsby" is the pursuit of the American Dream. The novel explores the idea that the American Dream is an illusion, and that the pursuit of wealth and success can lead to a life of emptiness and disillusionment.
& The central theme of "The Great Gatsby" is the pursuit of the American Dream. The novel explores the idea that the American Dream is an illusion, and that the pursuit of wealth and status can lead to moral decay and corruption.
&  "The Great Gatsby" explores the theme of the American Dream and the corrupting influence of wealth and materialism. The novel portrays the pursuit of the American Dream by Jay Gatsby, who uses his wealth to try to win back Daisy Buchanan, the woman he loved in his youth. However, the novel also shows the emptiness and superficiality of the lives of the wealthy characters, who are consumed by their pursuit of wealth and status. Ultimately, the novel suggests that the American Dream, which promises prosperity and happiness to all, is an illusion that can lead to moral decay and personal tragedy.
\\
\hline
\end{tabular}
}
\caption{Case Study in General Dialogue.}
\label{table:eff_general_dia}
\end{table*}

\begin{table*}[h!]
\centering \tiny
\resizebox{1.\linewidth}{!}{
\renewcommand{\arraystretch}{1.5}
\begin{tabular}{  m{3cm} | m{4cm} | m{6cm} }
\hline
\textbf{Questions} & \textbf{Mistral-Instruct} & \textbf{Mistral-C2F} \\
\hline
Scenario: A student is seeking clarification on a homework assignment posted on your e-learning platform. How would you provide assistance?

Student: "I don't understand question number 5 on the homework assignment. Can you explain it to me?"

& Of course, I'd be happy to help you understand question number 5 on your homework assignment. Before we dive into the specifics, could you please share which subject and topic the assignment is related to? Additionally, could you provide some context about what you've tried so far or any concepts that are giving you trouble? With this information, I can provide a more accurate and helpful explanation. If you're not comfortable sharing this information publicly, please feel free to send me a private message or email. I'm here to help!

& Of course, I'd be happy to help you understand question number 5 on the homework assignment. Let me first clarify the context of the question. The question is asking you to analyze the impact of a specific policy on a particular industry. To answer this question, you need to consider the following factors:

1. The nature of the policy: What is the policy trying to achieve? How does it affect the industry?

2. The industry's structure: What is the industry's size and composition? How does the policy impact different segments of the industry?

3. The policy's implementation: How is the policy being implemented? What are the challenges and opportunities associated with the implementation?

4. The policy's impact on stakeholders: Who are the stakeholders affected by the policy? How does the policy impact their interests and objectives?

By considering these factors, you should be able to analyze the impact of the policy on the industry. If you have any further questions or need additional clarification, please don't hesitate to ask. I'm here to help you succeed in your studies.
\\

\hline
You and a friend are disagreeing about a historical event you learned about in school. How would you frame a query to the model to get a balanced, fact-based explanation to settle the disagreement?
& I would frame the query as follows: "Model, could you please provide a fact-based explanation about [the specific historical event]? I'm having a disagreement with a friend and I'd like to ensure we both have accurate information. Please source your information from reputable academic or historical databases to ensure the highest level of factual accuracy."
& To get a balanced, fact-based explanation to settle a disagreement about a historical event, you can frame a query to the model in the following way:

"Please provide a balanced and fact-based explanation of the historical event [insert name of event here]. Include the main points of the event, the context in which it occurred, and any relevant facts or evidence. Please avoid taking sides or making judgments about the event, and instead focus on presenting the facts in an objective and unbiased manner."

By framing the query in this way, you are asking the model to provide a fact-based explanation of the event, without taking sides or making judgments. This should help you and your friend to better understand the event and come to a more informed and balanced conclusion.
\\

\hline
\end{tabular}
}
\caption{Case Study in General Dialogue.}
\label{table:eff_general_dia_instruct_and_rl}
\end{table*}

\subsection{Comprehensive Case Study on MT-Bench.}
\label{sec:more_mtbench_example}

\begin{table*}[h!]
\centering \tiny
\resizebox{1.\linewidth}{!}{
\renewcommand{\arraystretch}{1.5}
\begin{tabular}{  m{2.5cm} | m{3cm} | m{3cm} | m{5cm} }
\hline
\textbf{Questions} & \textbf{Mistral-Plus} & \textbf{Mistral-Plus-Fine} & \textbf{Mistral-C2F} \\
\hline
The city of Vega intends to build a bridge that will span the Vegona River, covering a distance of 1.8 kilometers. The proposed location falls within a seismically active area that has experienced several high-magnitude earthquakes. Given these circumstances, what would be the best approach to constructing the bridge?

& The best approach to constructing the bridge would be to use seismic-resistant materials and design the bridge to withstand the effects of earthquakes. Seismic-resistant materials include steel, reinforced concrete, and fiber-reinforced polymer. The bridge should be designed to withstand the lateral forces of an earthquake, including the potential for ground movement.

& Given the seismically active area, the best approach to constructing the bridge would be to use earthquake-resistant materials and design. This would include using reinforced concrete, steel, and other materials that can withstand the forces of an earthquake. Additionally, the bridge design should incorporate seismic isolation and energy dissipation systems to reduce the impact of seismic activity on the bridge.
&  1. Conduct a thorough geotechnical investigation of the site to determine the soil conditions and potential seismic hazards. This will help in selecting the appropriate foundation design and materials. \newline
2. Consider using a seismic isolation system for the bridge. This system involves isolating the bridge from the ground using flexible bearings or other isolation devices. This allows the bridge to move independently during an earthquake, reducing the forces transmitted to the structure and minimizing the potential for damage. \newline
3. Use high-strength, ductile materials for the bridge's superstructure, such as reinforced concrete or steel. These materials have the ability to deform and absorb energy during an earthquake, reducing the likelihood of structural failure. \newline
4. Implement strict quality control measures during the construction process to ensure that all materials and components meet the required specifications and are properly installed. \newline
5. Conduct regular inspections and maintenance of the bridge to identify and address any potential issues before they become major problems. \newline
6. Develop and implement an emergency response plan for the bridge in the event of an earthquake or other natural disaster. This plan should include procedures for evacuating the area, assessing the damage, and initiating repairs or replacement of the bridge as necessary. \newline
By following these steps, the city of Vega can construct a bridge that is resilient to seismic activity and can withstand the forces of even the strongest earthquakes.
\\
\hline
Thomas is very healthy, but he has to go to the hospital every day. What could be the reasons?

& 1. Thomas may have a chronic illness that requires regular medical attention. 2. Thomas may have an injury that requires ongoing treatment. 3. Thomas may have a mental health condition that requires regular medical care. 4. Thomas may have a genetic disorder that requires regular medical monitoring.

& 1. Thomas may have a chronic illness that requires daily hospital visits for treatment. 2. Thomas may be undergoing a medical procedure or treatment that requires daily hospital visits. 3. Thomas may be receiving palliative care or hospice care, which may require daily hospital visits. 4. Thomas may be a patient in a long-term care facility, which may require daily hospital visits for medical treatment.

& 1. Thomas may have a chronic illness that requires daily hospital visits for treatment or monitoring. 2. Thomas could be undergoing a medical procedure or therapy that requires daily hospital visits. 3. Thomas might be a hospital volunteer or employee who visits the hospital daily for work-related purposes. 4. Thomas could be a patient's companion or caregiver who accompanies the patient to the hospital daily. 5. Thomas might be a medical student or trainee who is required to visit the hospital daily for clinical rotations or training purposes. 6. Thomas could be a research subject who is participating in a clinical trial or study that requires daily hospital visits. 7. Thomas might be a patient who is undergoing a long-term hospitalization for a serious illness or injury. 8. Thomas could be a patient who is receiving palliative care or hospice services in a hospital setting. 9. Thomas might be a patient who is undergoing a medical procedure or therapy that requires daily hospital visits for monitoring or adjustments. 10. Thomas could be a patient who is receiving outpatient treatment or therapy at a hospital and is required to visit the hospital daily for these services.

\\
\hline
Help me construct a catchy, yet scientifically accurate, headline for an article on the latest discovery in renewable bio-energy, while carefully handling the ethical dilemmas surrounding bio-energy sources. Propose 4 options.
& 1. Bio-energy: The Future of Renewable Energy \newline
2. Bio-energy: A Sustainable Solution for the Future \newline 3. Bio-energy: A Promising Alternative to Fossil Fuels \newline 4. Bio-energy: A Clean and Efficient Energy Source
& 1. "Bio-Energy Breakthrough: Harnessing the Power of Nature"
2. "Ethical Bio-Energy: Balancing Progress and Responsibility"
3. "Renewable Bio-Energy: A Sustainable Solution for the Future"
4. "Bio-Energy Revolution: Transforming the Way We Power Our World"
& 1. "Revolutionary Bio-Energy Discovery: Harnessing Nature's Power with Ethical Considerations" \newline 2. "Bio-Energy Breakthrough: Ethically Sourced Renewable Energy for a Sustainable Future" \newline 3. "Ethically Sourced Bio-Energy: The Future of Renewable Energy with Scientific Accuracy" \newline 4. "Scientifically Accurate Bio-Energy Discovery: Ethical Considerations for a Sustainable Future" \newline These headlines aim to capture the reader's attention while accurately conveying the scientific nature of the discovery and the ethical considerations surrounding bio-energy sources. They also emphasize the potential for a sustainable future through the use of renewable bio-energy.
\\
\hline
\end{tabular}
}
\caption{Case Study in MT Bench.}
\label{table:eff_mt_app}
\end{table*}


\begin{table*}[h!]
\centering \tiny
\renewcommand{\arraystretch}{1.5}
\begin{tabular}{  m{3cm} | m{10cm} }
\hline
\textbf{Questions} & \textbf{Responses} \\
\hline
what ad policy does the following text violate ? This Streamline Makes Your Cheeky Hidden Hips Look Like a Juicy Peach & This text violates several Advertising Policy guidelines. First, the use of the word “cheeky” in the context of a product or service is likely to be considered misleading or deceptive by consumers. The use of the term “hidden hips” to describe a product or service is likely to be considered false or misleading, as it implies that the product or service can conceal or disguise the appearance of the hips. Additionally, the use of the term “juicy peach” to describe a product or service is likely to be considered inappropriate or offensive, as it is a sexualized and objectifying description. Furthermore, the use of the term “streamline” in the context of a product or service is likely to be considered misleading or deceptive, as it implies that the product or service can make the hips appear slimmer or more streamlined, which may not be the case. Overall, the use of these terms and phrases in the text is likely to violate several Advertising Policy guidelines and may be considered misleading or offensive to consumers. It is important for advertisers to use accurate and appropriate language in their advertising materials to avoid violating Advertising Policy guidelines and to ensure that their messages are clear, honest, and respectful to their target audience. If an advertiser is using the terms “cheeky,” “hidden hips,” “juicy peach,” and “streamline” in their advertising materials, it is important for them to ensure that these terms are used accurately and appropriately, and that they are not making any false or misleading claims about the product or service. Additionally, they should avoid using language that is sexualized or objectifying, and should ensure that their advertising materials are respectful and appropriate for their target audience. It may be helpful for the advertiser to seek the advice of a marketing or advertising professional to ensure that their advertising materials comply with Advertising Policy guidelines and are effective in communicating their message to their target audience. They may also consider conducting consumer research to understand how their target audience perceives the use of these terms and phrases in advertising materials, and to identify any potential concerns or objections that may arise. Ultimately...(exceed maximum length) \\
\hline
list all the prime numbers within 100 & The prime numbers within 100 are: 2, 3, 5, 7, 11, 13, 17, 19, 23, 29, 31, 37, 41, 43, 47, 53, 59, 61, 67, 71, 73, 79, 83, 89, 97 These are the prime numbers that are less than or equal to 100. A prime number is a natural number that is greater than 1 and has exactly two positive divisors, namely 1 and itself. In other words, a prime number is a number that can only be divided by itself and 1 without leaving a remainder. The prime numbers listed above are the first 31 prime numbers, starting with 2. It is worth noting that the number 1 is not considered a prime number because it has only one positive divisor, namely 1. Additionally, the number 2 is also not considered a prime number in the strict sense, as it has two positive divisors (1 and 2). However, it is often included in lists of prime numbers because it is the smallest prime number and plays a special role in number theory and other mathematical fields. Overall, the list of prime numbers within 100 provides a useful reference for various mathematical and computational applications. They can be used to test the primality of other numbers or to generate random prime numbers for encryption and other security-related purposes. Understanding prime numbers is also important for understanding the structure of the natural numbers and the properties of their divisors. In general, prime numbers are a fundamental concept in number theory and play a crucial role in various areas of mathematics and computer science. They are the building blocks of the natural numbers and have numerous applications in cryptographic algorithms, factorization methods, and other mathematical problems. Understanding prime numbers and their properties is essential for solving a wide range of mathematical and computational problems. Additionally, prime numbers have been the subject of intense study and research for centuries, and their properties and behavior have been the focus of many important mathematical theorems and conjectures. For example, the... (exceed maximum length)
\\
\hline
compose an article which starts with flowers. & Flowers are a beautiful and delicate part of the natural world that can bring joy and beauty to any space. They come in a wide variety of colors, shapes, and sizes, and can be found in many different environments around the world. From vibrant and bold blooms to delicate and fragrant blossoms, there is a flower to suit every taste and preference. Flowers are not only visually stunning, but they also have a number of benefits for both humans and the environment. For instance, they can help to purify the air, provide food and shelter for wildlife, and even have therapeutic properties. In many cultures, flowers are also used in ceremonies and celebrations to mark important occasions and express emotions. They can be given as gifts to loved ones to show affection, or used to decorate special events such as weddings and birthdays. In addition to their aesthetic and symbolic value, flowers can also be used in the culinary world. Many edible flowers have a unique flavor and can be used to add a burst of color and flavor to salads, desserts, and cocktails. Overall, flowers are a wonderful and versatile part of the natural world that can bring beauty, joy, and even practical benefits to our lives. Whether you enjoy growing them in your own garden, admiring them in public parks, or simply appreciating their beauty in floral arrangements, flowers are a delightful and uplifting part of the human experience. They remind us of the beauty and wonder of the natural world and the power of small, delicate things to bring us joy and inspiration. So the next time you come across a stunning bouquet of flowers or see a field of wildflowers in full bloom, take a moment to appreciate the beauty of these natural wonders and the many ways they enrich our lives. Flowers truly are a gift from nature that we should cherish and celebrate. I'm a little bit confused by the prompt. Is the task to write an article that starts with the word "flowers," or is it to compose an article about flowers? If it's the latter, then I would recommend starting with a general introduction that provides some background information on the topic of flowers. You could discuss the history of flower cultivation, the various types and species of flowers... (exceed maximum length)
\\
\hline
\end{tabular}
\caption{Case Study in Coarse Actor Generation.}
\label{table:eff_coarse_app}
\end{table*}

\begin{table*}[h!]
\centering \tiny
\renewcommand{\arraystretch}{1.5}
\begin{tabular}{ m{1.5cm} |  m{2.5cm} | m{10cm} }
\hline
\textbf{Model} & \textbf{Question} & \textbf{Response} \\
\hline
Coarse Actor 0.1 + Intruct 0.9 & compose an article which starts with flowers. &  Flowers are a symbol of beauty, life, and love. They have been a source of inspiration for poets, artists, and writers for centuries. The sight of a field of blooming flowers can bring a sense of peace and tranquility, while the scent of a rose can evoke memories of love and romance. Flowers are not only beautiful to look at, but they also play an important role in our ecosystem. They provide food and shelter for pollinators such as bees and butterflies, and they help to purify the air we breathe. In addition to their natural beauty and ecological importance, flowers have also been used throughout history for medicinal purposes. Many flowers contain compounds that have been shown to have anti-inflammatory, antioxidant, and anti-cancer properties. For example, chamomile flowers have been used to treat anxiety and insomnia, while turmeric flowers have been used to reduce inflammation and improve digestion. In conclusion, flowers are a source of beauty, inspiration, and healing. Whether you enjoy admiring them in a garden, using them to decorate your home, or incorporating them into your skincare routine, there is no denying the many benefits that flowers bring to our lives.

\\
\hline
Coarse Actor 0.2 + Intruct 0.8 & compose an article which starts with flowers. &   Title: The Language of Flowers: A Symphony of Silence

In the quietude of a serene morning, the garden comes alive with the gentle rustling of leaves and the sweet perfume of blossoms. Among these vibrant blooms, the language of flowers unfolds, a symphony of silence that speaks volumes.

The sun casts its golden rays upon the garden, illuminating the petals of the flowers, each one a unique masterpiece. The delicate pink hue of the rose, the vibrant orange of the marigold, the soft white of the lily, and the deep blue of the forget-me-not - each one tells a story, a tale of love, joy, sorrow, and hope.

The language of flowers, also known as floriography, is an ancient art that has been practiced for centuries. It is believed to have originated in China, where flowers were used to convey messages during the Han Dynasty. The art then spread to Japan, where it became an integral part of Japanese culture.
\\

\hline
Coarse Actor 0.3 + Intruct 0.7 & compose an article which starts with flowers. &   Title: The Language of Flowers: A Symphony of Silence

The flower, in its simple beauty, is a testament to the power of nature to inspire and uplift the human spirit. From the delicate petals of the rose to the bold, vibrant hues of the sunflower, each flower holds a unique and beautiful story.

One of the most famous and enduring examples of the use of the language of flowers in Chinese culture is the story of the "Legend of the White Snake." This classic Chinese folktale, which has been adapted into various forms of Chinese art and literature, including opera, poetry, and painting, tells the story of a white snake spirit who transforms herself into a beautiful young woman in order to fall in love with a mortal man.

Another example of the use of the language of flowers in the "Legend of the White Snake" can be found in a scene where the white snake spirit uses the language of flowers to communicate a warning to the mortal man she loves, as well as to other characters in the story. 

Today, the language of flowers continues to be a source of inspiration and fascination for people all around the world. From the simple pleasure of arranging a bouquet of flowers for a loved one, to the more complex and nuanced use of the language of flowers in various forms of art, literature, and culture, the beauty and richness of the language of flowers continues to captivate and inspire the imagination of people all around the world.
\\

\hline
Coarse Actor 0.4 + Intruct 0.6 & compose an article which starts with flowers. &   Title: The Language of Flowers: A Symphony of Silent Communication

Flowers have been a source of beauty and inspiration for centuries. They are not just pretty decorations, but also carriers of silent messages.

The ancient Greeks and Romans believed that flowers held mystical powers and could convey hidden meanings. This belief was further popularized in the Victorian era, when the art of sending coded messages through flowers, known as floriography, became a popular pastime.

Each flower was assigned a specific meaning, and by combining different flowers, one could convey a complex message. For example, a bouquet of roses, lilies, and violets could be interpreted as a declaration of love, loyalty, and trust.

Today, the art of sending messages through flowers may not be as popular as it once was, but the language of flowers continues to inspire and captivate us. Whether it's the delicate scent of a rose, the vibrant colors of a sunflower, or the soothing touch of a daisy, flowers have the power to evoke emotions and convey messages in a way that words alone cannot.

So the next time you receive a bouquet of flowers, take a moment to appreciate the silent messages they carry. Whether it's a declaration of love, a gesture of sympathy, or a simple expression of gratitude, the language of flowers is a beautiful and enduring testament to the power of communication.
\\

\hline
Coarse Actor 0.5 + Intruct 0.5 & compose an article which starts with flowers.
&   Title: The Language of Flowers: A Symphony of Silent Communication

Flowers have been a source of beauty and inspiration for centuries. They are not just pretty decorations, but also powerful symbols that convey a wide range of emotions and messages.

The language of flowers, also known as floriography, is an ancient art that dates back to the Victorian era. During this time, people used flowers to express feelings that were considered too sensitive or inappropriate to be spoken aloud.

Each flower has its own unique meaning, and when arranged in a bouquet, they can tell a story that is as complex and nuanced as human emotions themselves.

For example, a bouquet of roses can convey a multitude of meanings, depending on the color of the roses. Red roses symbolize love and passion, while white roses represent purity and innocence. Pink roses, on the other hand, signify gratitude and appreciation.

Other popular flowers and their meanings include:

- Lilies: symbolize purity and innocence
- Tulips: represent perfect love
- Daisies: signify innocence and purity
- Sunflowers: represent loyalty and adoration
- Orchids: symbolize luxury and refinement

The language of flowers is a beautiful and intricate art that has been passed down through the generations. It is a testament to the power of nature to inspire and uplift us, and to the enduring human desire to find meaning and connection in the world around us.\\
\hline
\end{tabular}
\end{table*}

\begin{table*}[h!]
\centering \tiny
\renewcommand{\arraystretch}{1.5}
\begin{tabular}{ m{1.5cm} |  m{2.5cm} | m{10cm} }
\hline
\textbf{Model} & \textbf{Question} & \textbf{Response} \\
\hline
Coarse Actor 0.6 + Intruct 0.4 & compose an article which starts with flowers. &  Title: The Beauty of Flowers: A Symphony of Colors and Fragrances

Flowers have been a source of inspiration and delight for centuries. From the ancient Egyptians who used flowers in their burial rituals, to the Romantic poets who wrote about the beauty of flowers, these natural wonders have captured the imagination of people all over the world.

The first thing that comes to mind when we think of flowers is their stunning beauty. With their vibrant colors and delicate petals, flowers are a feast for the eyes. Whether it's the bright red of a rose, the soft pink of a peony, or the cheerful yellow of a sunflower, each flower has its own unique beauty that never fails to captivate us.

But flowers are not just beautiful to look at. They also have a wonderful fragrance that can transport us to another world. From the sweet scent of a jasmine flower to the earthy aroma of a lily, each flower has its own distinctive fragrance that can evoke powerful emotions and memories.

Flowers are also an important part of our culture and traditions. In many countries, flowers are used to express emotions and sentiments. For example, in the language of flowers, a red rose is a symbol of love and passion, while a white rose represents purity and innocence.

In addition to their cultural significance, flowers also have practical uses. Many flowers are used in medicine to treat a variety of ailments. For example, chamomile flowers are often used to make tea that can help soothe an upset stomach or promote better sleep.

In conclusion, flowers are truly a wonder of nature. With their stunning beauty, wonderful fragrance, cultural significance, and practical uses, flowers are a source of inspiration and delight for people all over the world. So the next time you see a beautiful flower, take a moment to appreciate its beauty and wonder.

\\
\hline
Coarse Actor 0.7 + Intruct 0.3 & compose an article which starts with flowers. &  Title: The Language of Flowers: A Symphony of Silence

Flowers, the epitome of beauty and grace, have long been a source of inspiration and wonder. From the delicate petals of a rose to the vibrant hues of a sunflower, flowers have the power to evoke emotions and convey messages that words alone cannot express.

The ancient Greeks believed that flowers were the embodiment of the gods and goddesses. The rose, for instance, was associated with Aphrodite, the goddess of love and beauty. The lotus flower, on the other hand, was sacred to the sun god, Helios.

In many cultures, flowers have been used as symbols of various emotions and sentiments. The red rose, for example, is a universal symbol of love and passion. The white rose, on the other hand, represents purity and innocence.

The language of flowers, also known as floriography, was a popular form of communication during the Victorian era. Through the use of specific flowers and their meanings, people were able to convey complex emotions and messages that could not be expressed through words alone.

In conclusion, flowers have long been a source of inspiration and wonder. From their symbolic meanings to their ability to convey complex emotions and messages, flowers have the power to evoke emotions and connect people in ways that words alone cannot express. So the next time you come across a beautiful flower, take a moment to appreciate its beauty and the message it may be trying to convey.
\\

\hline
Coarse Actor 0.8 + Intruct 0.2 & compose an article which starts with flowers. &   Flowers are a beautiful and beloved part of nature. They come in a wide variety of colors, shapes, and sizes, and they have been admired and appreciated by humans for centuries. Flowers have a special ability to brighten up any space and bring a sense of joy and happiness to those who see them. They are often used as symbols of love, friendship, and celebration, and they are a popular choice for gifts and decorations. Flowers also play an important role in the ecosystem, providing food and shelter for a variety of insects and other animals. They help to purify the air and contribute to the overall health and well-being of the environment. In addition to their aesthetic and ecological value, flowers also have a long history of use in traditional medicine. Many flowers contain compounds that have been shown to have therapeutic properties, and they have been used to treat a wide range of ailments, from headaches and colds to more serious conditions like cancer and heart disease. Overall, flowers are a truly remarkable and multifaceted part of our world. They bring beauty and joy to our lives, contribute to the health and well-being of the environment, and have a rich history of use in traditional medicine. Whether you're admiring a bouquet of freshly cut flowers, or simply enjoying the sight of a beautiful flowering plant in your garden, there's no denying the power and magic of these wonderful creations of nature.
\\

\hline
Coarse Actor 0.9 + Intruct 0.1 & compose an article which starts with flowers. &   Flowers are a beautiful and beloved part of nature, bringing color and life to our surroundings. From the delicate petals of a rose to the vibrant hues of a sunflower, flowers have the power to brighten our mood and bring joy to our lives. But flowers are not just pretty to look at – they also have a rich history and cultural significance. In many cultures, flowers are used to express emotions and sentiments, with different types of flowers representing different meanings. For example, roses are often associated with love and romance, while lilies are seen as symbols of purity and innocence. Flowers are also used in various ceremonies and celebrations, such as weddings, funerals, and graduations. They are often given as gifts to show appreciation, sympathy, or congratulations. In addition to their aesthetic and cultural value, flowers also play an important role in the environment. They are essential for pollination, which is the process by which pollen is transferred from one plant to another, allowing for the reproduction of plants and the continuation of life. Flowers also provide food and shelter for a wide range of insects and other animals, making them an important part of the food chain. Overall, flowers are a fascinating and multifaceted part of our world, with a rich history and cultural significance, as well as important environmental and ecological functions. Whether you're admiring them in a garden, giving them as a gift, or using them in a ceremony, flowers are a reminder of the beauty and wonder of nature, and the importance of appreciating the small things in life. Flowers are a symbol of beauty, fragility, and rebirth. They have been used throughout history to express emotions, celebrate special occasions, and adorn our homes and gardens. The beauty of flowers lies in their delicate petals, vibrant colors, and sweet scents. They come in a wide variety of shapes, sizes, and colors, each with its own unique characteristics and symbolism. For example, roses are often associated with love and romance, while lilies are seen as symbols of purity and innocence. Flowers are also used to express emotions and sentiments, with different types of flowers representing different meanings. For instance, sunflowers are often given as a symbol of loyalty and devotion, while daisies are seen as a symbol of innocence and purity. Flowers are also used in various ceremonies and celebrations, such as weddings, funerals, and graduations. They are often given as gifts to show appreciation, sympathy, or congratulations. In addition to their aesthetic and cultural value, flowers also play an important role in the environment. They are essential for pollination, which is the process by which pollen is transferred from one plant to another, allowing for the reproduction of plants and the continuation of life. Flowers also provide food and shelter for a wide range of insects and other animals, making them an important part of the food chain. Overall, flowers are a fascinating and multifaceted part of our world, with a rich history and cultural significance, as well as important environmental and ecological functions. Whether you're admiring them in a garden, giving them as a gift, or using them in a ceremony, flowers are a reminder of the beauty and wonder of nature, and the importance of appreciating the small things in life.
\\

\hline
\end{tabular}
\caption{Case Study in Fine Actor Generation task between difference ratios.}
\label{table:eff_fine_2}
\end{table*}
